\pdfoutput=1

\documentclass[11pt]{article}

\usepackage[]{acl}

\usepackage{times}
\usepackage{latexsym}

\usepackage[T1]{fontenc}

\usepackage[utf8]{inputenc}

\usepackage{microtype}

\usepackage{inconsolata}

\usepackage{amsmath}
\usepackage{graphicx}
\usepackage{booktabs}
\usepackage{arydshln}
\usepackage{tabularx}
\usepackage{amsfonts}
\usepackage{multirow}
\usepackage{multicol}
\usepackage{cleveref}
\usepackage{xspace}
\usepackage{makecell}
\usepackage{fontawesome5}
\usepackage{colortbl}
\usepackage[ruled, lined, linesnumbered, commentsnumbered]{algorithm2e}
\usepackage{todonotes}
\usepackage{tablefootnote}
\usepackage{tabularray}
\usepackage{xspace}
\usepackage{circledsteps}
\usepackage{caption}
\usepackage{longtable}
\usepackage{multicol}
\usepackage{array}

\usepackage{listings}
\definecolor{dkgreen}{rgb}{0,0.6,0}
\definecolor{gray}{rgb}{0.5,0.5,0.5}
\definecolor{mauve}{rgb}{0.58,0,0.82}
\lstdefinelanguage{Python}{
 keywords={and, as, assert, break, class, continue, def, del, elif, else,
 except, False, finally, for, from, global, if, import, in, is, lambda,
 None, nonlocal, not, or, pass, raise, return, True, try, while, with, yield, str, dspy},
 keywordstyle=\color{blue},
 emph={format}, %
 emphstyle={\color{black}}, %
 stringstyle=\color{dkgreen}, %
 commentstyle=\color{dkgreen}, %
 morecomment=[s]{"""}{"""},
 morestring=[b]',
 morestring=[s]{'''}{'''},
}

\newcommand{\system}[0]{STORM\xspace}

\crefformat{section}{\S#2#1#3} %
\crefformat{subsection}{\S#2#1#3}
\crefformat{subsubsection}{\S#2#1#3}
\newcommand{\refequ}[1]{Equation~(\ref{#1})}
\newcommand{\reffig}[1]{Figure~\ref{#1}}
\newcommand{\refsec}[1]{\S\ref{#1}} %
\newcommand{\reftab}[1]{Table~\ref{#1}}

\def\eg{\textit{e.g.}\xspace}

\def\etc{\textit{etc.}\xspace}
\def\ie{\textit{i.e.}\xspace}

\newcommand{\RNum}[1]{\uppercase\expandafter{\romannumeral #1\relax}}

\definecolor{myyellow}{RGB}{249, 231, 173}
\definecolor{mygreen}{RGB}{233, 242, 230}

\title{Assisting in Writing Wikipedia-like Articles From Scratch \\ with Large Language Models}

\author{Yijia Shao\quad Yucheng Jiang\quad Theodore A. Kanell\quad Peter Xu\\\textbf{Omar Khattab}\quad \textbf{Monica S. Lam}\\
Stanford University\\
\texttt{\{shaoyj, yuchengj, tkanell, peterxu, okhattab\}@stanford.edu}\\
\texttt{lam@cs.stanford.edu}}

\begin{document}
\maketitle
\begin{abstract}

We study how to apply large language models to write grounded and organized long-form articles from scratch, with comparable breadth and depth to Wikipedia pages. This underexplored problem poses new challenges at the {\em pre-writing} stage, including how to research the topic and prepare an outline prior to writing.
We propose \textbf{\system}, a writing system for the \textbf{S}ynthesis of \textbf{T}opic \textbf{O}utlines through \textbf{R}etrieval and \textbf{M}ulti-perspective Question Asking.
\system models the pre-writing stage by %
(1) discovering diverse perspectives in researching the given topic, 
(2) simulating conversations where writers carrying different perspectives pose %
questions to a topic expert grounded on trusted Internet sources, (3) curating the collected information to create an outline. 

For evaluation, we curate FreshWiki, a dataset of recent high-quality Wikipedia articles, and formulate outline assessments to evaluate the pre-writing stage. 
We further gather feedback %
from experienced Wikipedia editors. Compared to articles generated by an outline-driven retrieval-augmented baseline, more of STORM's articles are deemed to be organized (by a 25\% absolute increase) and broad in coverage (by 10\%).
The expert feedback also helps identify new challenges for generating grounded long articles, such as source bias transfer and over-association of unrelated facts.

\end{abstract}

\section{Introduction}
\label{sec:intro}
\begin{figure}[t]
    \centering 
    \resizebox{0.95\columnwidth}{!}{%
    \includegraphics{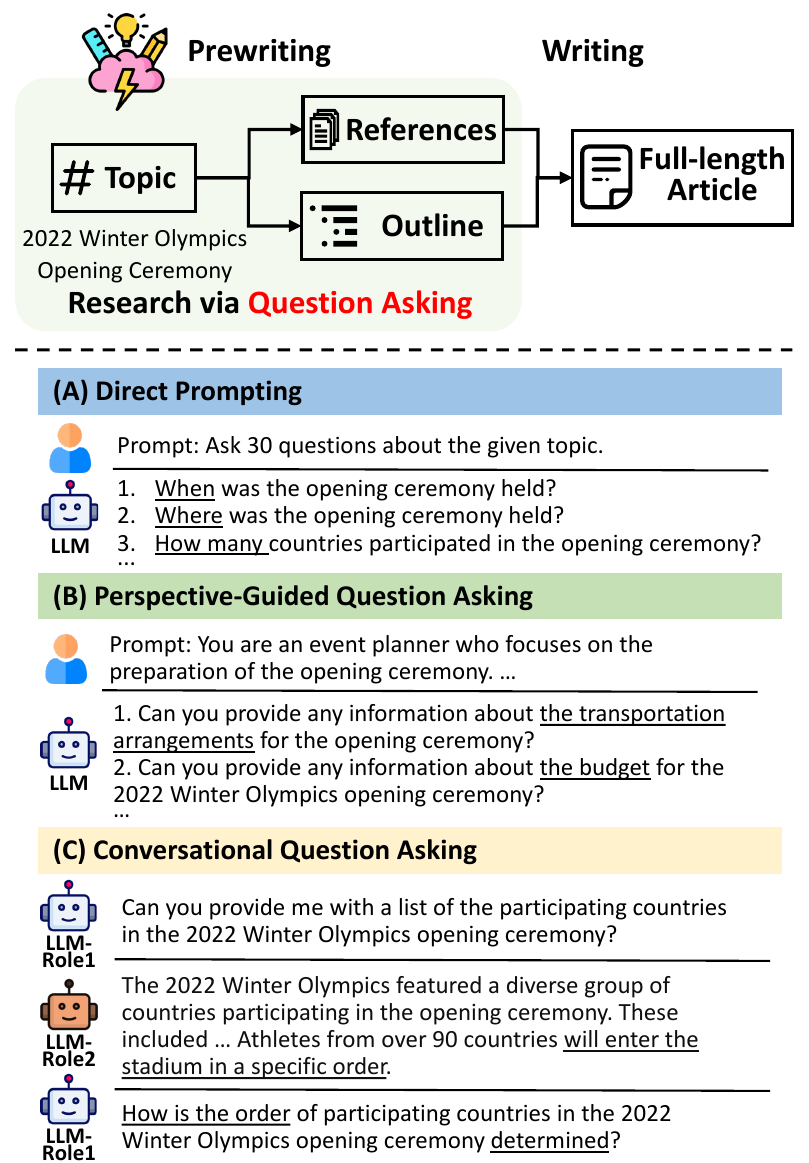}
    }
    \caption{We explore writing Wikipedia-like articles from scratch, which demands a pre-writing stage before producing the article. In this stage, simpler approaches like Direct Prompting have limited planning capacity. In contrast, \system researches the topic via perspective-guided question asking in simulated conversations.%
    }
    \label{Fig.question_asking}
\vspace{-0.5em}
\end{figure}

Large language models (LLMs) have demonstrated impressive writing capabilities~\citep{yang-etal-2023-doc, pavlik2023collaborating,wenzlaff2022smarter,fitria2023artificial}, but it is unclear how we can use them to write grounded, long-form articles, like full-length Wikipedia pages. Such expository writing, which seeks to inform the reader on a topic in an organized manner~\citep{weaver1991expository, balepur-etal-2023-expository}, requires thorough research and planning in the \textit{pre-writing} stage~\citep{rohman1965pre}, even before the actual writing process can start.
However, prior work on generating Wikipedia articles~\citep{banerjee-mitra-2015-wikikreator,minguillon2017semi,j.2018generating,fan-gardent-2022-generating} has generally bypassed the pre-writing stage: for instance, \citet{j.2018generating} presume reference documents are provided in advance, while \citet{fan-gardent-2022-generating} assume an article outline is available and focus on expanding each section. These assumptions do not hold in general, as collecting references and crafting outlines demand advanced information literacy skills~\citep{doyle1994information} to identify, evaluate, and organize external sources - a task that is challenging even for experienced writers. Automating this process can facilitate individuals in initiating in-depth learning about a topic and greatly reduce the expensive expert hours necessary for their expository writing.

We explore these challenges by focusing on how to generate Wikipedia-like articles \textit{from scratch}. %
We decompose this problem into two tasks. The first is to conduct research to generate an outline, \ie, a list of multi-level sections, and collect a set of reference documents. The second uses the outline and the references to produce the full-length article. Such a task decomposition mirrors the human writing process which usually includes phases of pre-writing, drafting, and revising~\citep{rohman1965pre,munoz2015main}.

As pre-trained language models inherently possess a wealth of knowledge, a direct approach is to rely on their parametric knowledge for generating outlines or even entire articles (\textit{Direct Gen}). However, this approach is limited by a lack of details and hallucinations~\citep{xu-etal-2023-critical}, particularly in addressing long-tail topics~\cite{kandpal2023large}. This underscores the importance of leveraging external sources, and current strategies often involve retrieval-augmented generation (\textit{RAG}), which circles back to the problem of researching the topic in the pre-writing stage, as much information cannot be surfaced through simple topic searches.

Human learning theories~\citep{tawfik2020role,booth2003craft}  highlight \textit{asking effective questions} in information acquisition.  
Although instruction-tuned models~\citep{ouyang2022training} can be prompted directly to generate questions, we find that they typically produce basic ``What'', ``When'', and ``Where'' questions (\reffig{Fig.question_asking} (A)) which often only address surface-level facts about the topic. 
To endow LLMs with the capacity to conduct better research, we propose the \textbf{\system} paradigm for the \textbf{S}ynthesis of \textbf{T}opic \textbf{O}utlines through \textbf{R}etrieval and \textbf{M}ulti-perspective Question Asking.

The design of \system is based on two hypotheses: (1) diverse perspectives lead to varied questions; (2) formulating in-depth questions requires iterative research. Building upon these hypotheses, \system employs 
a novel multi-stage approach. It first discovers diverse perspectives by retrieving and analyzing Wikipedia articles from similar topics and then personifies the LLM with specific perspectives for question asking (\reffig{Fig.question_asking} (B)). Next, to elicit follow-up questions for iterative research (\reffig{Fig.question_asking} (C)), %
\system simulates multi-turn conversations where the answers to the generated questions are grounded on the Internet. Finally, based on the LLM's internal knowledge and the collected information, \system creates an outline that can be expanded section by section to develop a full-length Wikipedia-like article.

We evaluate \system using our \textbf{FreshWiki} dataset (\refsec{sec:task_dataset}) which curates recent, high-quality Wikipedia articles to avoid data leakage during pre-training.\footnote{Our resources and code are released at \url{https://github.com/stanford-oval/storm}.}
To facilitate the study of the pre-writing stage, we define metrics for evaluating the outline quality against human-written articles. 

We further invited a group of experienced Wikipedia editors for expert evaluation. %
The editors found STORM outperforms an outline-driven RAG baseline, especially regarding the breadth and organization of the articles. They also identified %
challenges for future research, including addressing cases where:
(1) the bias %
on the Internet affects the generated articles; (2) LLMs fabricate connections between unrelated facts. 
These challenges present new frontiers to grounded writing systems.

Our main contributions include:
\begin{itemize}
\item {To evaluate the capacity of LLM systems at generating long-form grounded articles from scratch, and the pre-writing challenge in particular, we curate the FreshWiki dataset and establish evaluation criteria for both outline and final article quality.}
\item {We propose \system, a novel system that automates the pre-writing stage. \system researches the topic and creates an outline by using LLMs to ask incisive questions and retrieving trusted information from the Internet.}
\item{Both automatic and human evaluation demonstrate the effectiveness of our approach. Expert feedback further reveals new challenges in generating grounded long-form articles.}

\end{itemize}

\section{FreshWiki}
\label{sec:task}

We study generating Wikipedia-like articles from scratch, placing emphasis on the \textit{pre-writing} stage~\citep{rohman1965pre}, which involves the demanding sub-tasks of gathering and curating relevant information (``research''). %
This models the human writing approach which has prompted
some educators to view %
Wikipedia article writing 
as an educational exercise for academic training%
~\citep{tardy2010writing}.

\begin{table}
\centering
\resizebox{\columnwidth}{!}{%
\begin{tabular}{lcccc} 
\toprule
     & \textbf{Domain} & \textbf{Scope} & \begin{tabular}[c]{@{}c@{}}\textbf{Given}\\\textbf{Outline?}\end{tabular} & \begin{tabular}[c]{@{}c@{}}\textbf{Given}\\\textbf{Refs?}\end{tabular}  \\ 
\midrule
  \citet{balepur-etal-2023-expository}   & One             & One para.      & /                                                                         & Yes                                                                     \\
 \citet{qian2023webbrain}    & All             & One para.      & /                                                                         & No                                                                      \\
  \citet{fan-gardent-2022-generating}   & One             & Full article   & Yes                                                                       & No                                                                      \\
 \citet{j.2018generating}    & All             & One para.      & /                                                                         & Yes                                                                     \\
 \citet{sauper-barzilay-2009-automatically}    & Two             & Full article   & No                                                                        & No                                                                      \\ 
\midrule
Ours & All             & Full article   & No                                                                        & No                                                                      \\
\bottomrule
\end{tabular}
}
\caption{Comparison of different Wikipedia generation setups in existing literature. Generating one paragraph does not need an article outline.}
\vspace{-0.5em}
\label{table:background}
\end{table}

\reftab{table:background} compares our work against prior benchmarks for Wikipedia generation. Existing work has generally focused on evaluating the generation of shorter snippets (\eg, one paragraph), within a narrower scope (\eg, a specific domain or two), or when an explicit outline or reference documents are supplied. A notable example is WikiSum~\citep{j.2018generating}, which treats generating Wikipedia articles as a multi-document summarization problem, with respect to the reference documents. 

Our setup emphasizes the capability of long-form grounded writing systems to research and curate content. Specifically, given a topic $t$, the task is to find a set of references $\mathcal{R}$ and generate a full-length article $\mathcal{S}=s_1s_2...s_n$, where each sentence $s_i$ cites a list of documents in $\mathcal{R}$.\footnote{In practice, $\mathcal{S}$ also includes organizational elements such as section and subsection titles, which do not require citations.}

\subsection{The FreshWiki Dataset}
\label{sec:task_dataset}

Creating a new Wikipedia-like article demands not only fluent writing but also good research skills. As modern LLMs are generally trained on Wikipedia text, we mitigate data leakage by explicitly seeking out \textit{recent} Wikipedia articles that were created (or very heavily edited) after the training cutoff of the LLMs we test. Our process can be repeated at future dates when new LLMs emerge.

To apply our date criteria, we focus on the top 100 most-edited pages, based on edit counts, for each month from February 2022 to September 2023\footnote{Obtained from \url{\detokenize{https://wikimedia.org/api/rest_v1/metrics/edited-pages/top-by-edits/en.wikipedia/all-editor-types/content/}}\{year\}/\{month\}/all-days}. 
To ensure high-quality references, we filter these articles to keep only those having B-class quality or above assessed by ORES\footnote{\url{https://www.mediawiki.org/wiki/ORES}}. We also exclude list articles\footnote{\url{https://en.wikipedia.org/wiki/Wikipedia:Stand-alone_lists}} and articles that have no subsections. While high-quality Wikipedia articles usually contain structured data (\eg, tables) and are multi-modal, we only consider the plain text component in constructing the dataset to simplify our task. More details of the dataset are in Appendix~\ref{appendix:dataset}.

\subsection{Outline Creation and Evaluation}
\label{sec:task_outline}

A full-length article is hard to generate or evaluate~\citep{xu-etal-2023-critical, krishna-etal-2023-longeval}. When human educators teach students academic writing, they sometimes supervise students at the outline stage~\citep{eriksson2015supervision} because an extensive outline indicates a comprehensive understanding of the topic and provides a solid foundation for writing the full-length article~\citep{dietz2019trec}. Inspired by this, we decompose the generation of $\mathcal{S}$ into two stages. In the pre-writing stage, we require the system to create an outline $\mathcal{O}$, which is defined as a list of multi-level section headings\footnote{Since language models process and produce sequences, we can linearize $\mathcal{O}$ by adding ``\#'' to indicate section titles, ``\#\#'' to indicate subsection titles, etc.}. In the writing stage, the system uses the topic $t$, the references $\mathcal{R}$, and an outline $\mathcal{O}$ to produce the full-length article $\mathcal{S}$.

To evaluate the outline coverage, we introduce two metrics: \textit{heading soft recall} and \textit{heading entity recall}. These metrics compare the multi-level section headings of the human-written article, considered as ground truth, and those in $\mathcal{O}$. Recognizing that an exact match between elements in these two sets of headings is unnecessary, we calculate the \textit{heading soft recall}~\citep{franti2023soft} using cosine similarity derived from Sentence-BERT~\citep{reimers-gurevych-2019-sentence} embeddings of the headings (details in Appendix~\ref{appendix:soft_heading_recall}).
We also compute the \textit{heading entity recall} which is quantified as the percentage of named entities in human-written article headings covered by $\mathcal{O}$. We extract entities with FLAIR named entity recognition (NER)~\citep{akbik-etal-2019-flair}.

\section{Method}
\label{sec:method}
\begin{figure*}[t]
    \centering 
    \resizebox{0.95\textwidth}{!}{%
    \includegraphics{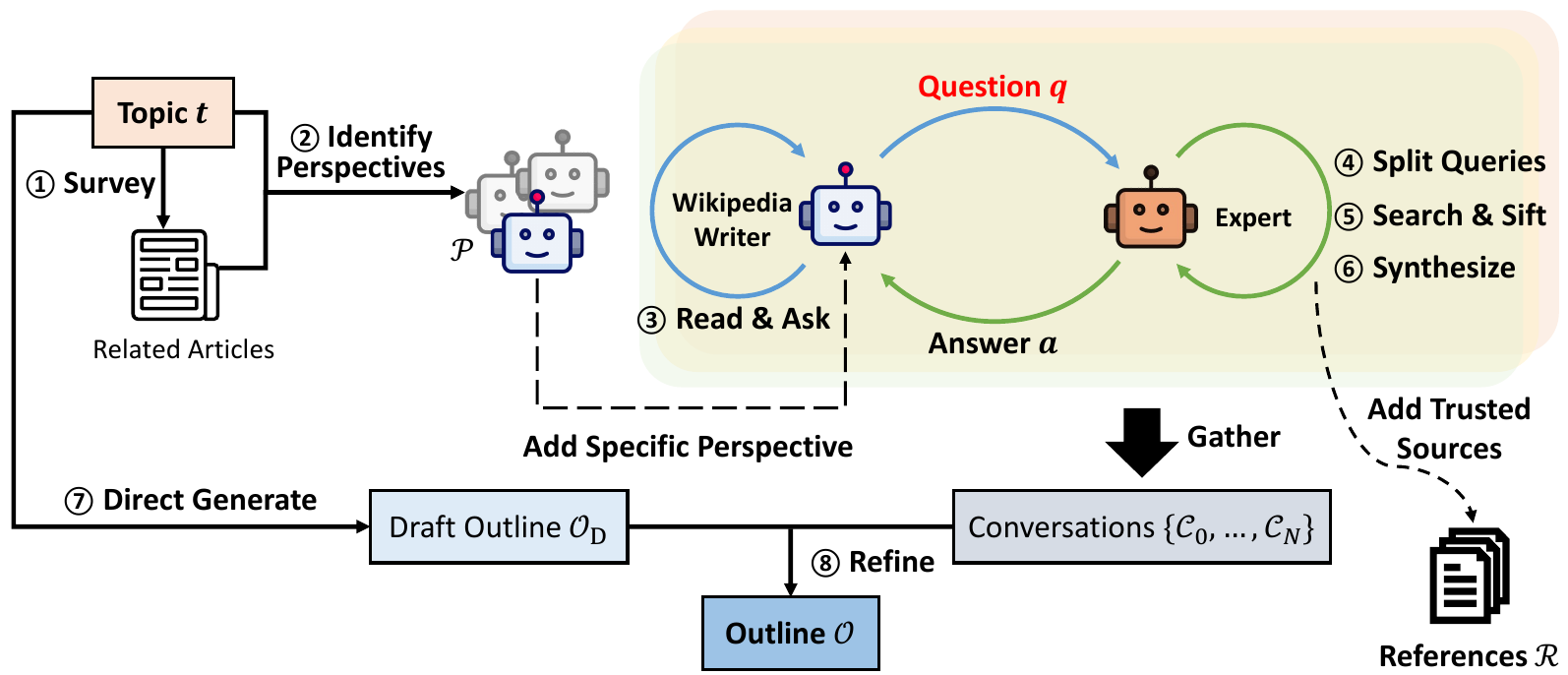}
    }
    \caption{The overview of \system that automates the pre-writing stage. Starting with a given topic, \system identifies various perspectives on covering the topic by surveying related Wikipedia articles (\Circled{1}-\Circled{2}). It then simulates conversations between a Wikipedia writer who asks questions guided by the given perspective and an expert grounded on trustworthy online sources (\Circled{3}-\Circled{6}). The final outline is curated based on the LLM's intrinsic knowledge and the gathered conversations from different perspectives (\Circled{7}-\Circled{8}).}
    \label{Fig.method}
\vspace{-0.5em}
\end{figure*}

We present \system to automate the pre-writing stage by researching a given topic via effective question asking (\refsec{sec:method_perspective},~\refsec{sec:method_conv}) and creating an outline (\refsec{sec:method_outline}). The outline will be extended to a full-length article grounded on the collected references (\refsec{sec:method_article}).~\reffig{Fig.method} gives an overview of \system and we include the pseudo code in Appendix~\ref{appendix:pseudo_code}.

\subsection{Perspective-Guided Question Asking}
\label{sec:method_perspective}
\citet{rohman1965pre} defines pre-writing as the stage of discovery in the writing process. 
In parallel with stakeholder theory in business~\citep{freeman2010stakeholder}, where diverse stakeholders prioritize varying facets of a company, individuals with distinct perspectives may concentrate on different aspects when researching the same topic and discover multifaceted information. Further, the specific perspectives can serve as prior knowledge, guiding individuals to ask more in-depth questions. %
For example, an event planner might ask about the ``transportation arrangements'' and ``budget'' for ``the 2022 Winter Olympics opening ceremony'', whereas a layperson might ask more general questions about the event's basic information (\reffig{Fig.question_asking} (A)).

Given the input topic $t$, \system discovers different perspectives by surveying existing articles from similar topics and uses these perspectives to control the question asking process. Specifically, \system prompts an LLM to generate a list of related topics and subsequently extracts the tables of contents from their corresponding Wikipedia articles, if such articles can be obtained through Wikipedia API\footnote{\url{https://pypi.org/project/Wikipedia-API/}} (\reffig{Fig.method}~\Circled{1}). These tables of contents are concatenated to create a context to prompt the LLM to identify $N$ perspectives $\mathcal{P}=\{p_1,...,p_N\}$ that can collectively contribute to a comprehensive article on $t$ (\reffig{Fig.method}~\Circled{2}). To ensure that the basic information about $t$ is also covered, we add $p_0$ as ``basic fact writer focusing on broadly covering the basic facts about the topic'' into $\mathcal{P}$. Each perspective $p \in\mathcal{P}$ will be utilized to guide the LLM in the process of question asking in parallel. 

\subsection{Simulating Conversations}
\label{sec:method_conv}
The theory of questions and question asking~\citep{ram1991theory} highlights that while answers to existing questions contribute to a more comprehensive understanding of a topic, they often simultaneously give rise to new questions. To kick off this dynamic process, \system simulates a conversation between a Wikipedia writer and a topic expert. In the $i$-th round of the conversation, the LLM-powered Wikipedia writer generates a single question $q_i$ based on the topic $t$, its assigned perspective $p\in\mathcal{P}$, and the conversation history $\{q_1, a_1,...,q_{i-1}, a_{i-1}\}$ where $a_j$ denotes the simulated expert's answer. The conversation history enables the LLM to update its understanding of the topic and ask follow-up questions. In practice, we limit the conversation to at most $M$ rounds.

To ensure that the conversation history provides factual information, we use trusted sources from the Internet to ground the answer $a_i$ to each query $q_i$. 
Since $q_i$ can be complicated, we first prompt the LLM to break down $q_i$ into a set of search queries (\reffig{Fig.method}~\Circled{4}) and the searched results will be evaluated using a rule-based filter according to the Wikipedia guideline\footnote{\url{https://en.wikipedia.org/wiki/Wikipedia:Reliable_sources}} to exclude untrustworthy sources (\reffig{Fig.method}~\Circled{5}). Finally, the LLM synthesizes the trustworthy sources to generate the answer $a_i$, and these sources will also be added to $\mathcal{R}$ for full article generation (\refsec{sec:method_article}).

\subsection{Creating the Article Outline}
\label{sec:method_outline}
After thoroughly researching the topic through $N+1$ simulated conversations, denoted as $\{\mathcal{C}_0, \mathcal{C}_1,..., \mathcal{C}_N\}$, \system creates an outline before the actual writing starts. To fully leverage the internal knowledge of LLMs, we first prompt the model to generate a draft outline $\mathcal{O}_\text{D}$ given only the topic $t$ (\reffig{Fig.method}~\Circled{7}). $\mathcal{O}_\text{D}$ typically provides a general but organized framework. Subsequently, the LLM is prompted with the topic $t$,  the draft outline $\mathcal{O}_\text{D}$, and the simulated conversations $\{\mathcal{C}_0, \mathcal{C}_1,..., \mathcal{C}_N\}$ to refine the outline (\reffig{Fig.method}~\Circled{8}). This results in an improved outline $\mathcal{O}$ which will be used for producing the full-length article.

\subsection{Writing the Full-Length Article}
\label{sec:method_article}

Building upon the references $\mathcal{R}$ collected and the outline $\mathcal{O}$ developed during the pre-writing stage, the full-length article can be composed section by section. Since it is usually impossible to fit the entire $\mathcal{R}$ within the context window of the LLM, we use the section title and headings of its all-level subsections to retrieve relevant documents from $\mathcal{R}$ based on semantic similarity calculated from Sentence-BERT embeddings. With the relevant information at hand, the LLM is then prompted to generate the section with citations. Once all sections are generated, %
they are concatenated to form the full-length article. %
Since the sections are generated in parallel, we prompt the LLM with the concatenated article to delete repeated information to improve coherence. Furthermore, in alignment with Wikipedia's stylistic norms, the LLM is also utilized to synthesize a summary of the entire article, forming the lead section at the beginning.

\section{Experiments}
\label{sec:exp}
\subsection{Article Selection}
\system is capable of researching complicated topics and writing long articles from detailed outlines. However, in this controlled experiment, we limit the final output to at most 4000 tokens (roughly 3000 words). For a meaningful comparison, we randomly select 100 samples from the FreshWiki dataset (see~\refsec{sec:task_dataset}) that have human-written articles not exceeding 3000 words.%

\subsection{Automatic Metrics}
\label{sec:automatic_metrics}
As discussed in~\refsec{sec:task_outline}, we evaluate the outline quality %
to assess the pre-writing stage by calculating the \textit{heading soft recall} and \textit{heading entity recall}. %
A higher recall score signifies a more comprehensive outline relative to the human-written article.

To assess the full-length article quality, %
we adopt \textit{ROUGE scores}~\citep{lin-2004-rouge} and compute the \textit{entity recall} in the article level based on FLAIR NER results. %
Moreover, based on Wikipedia criteria\footnote{\url{https://en.wikipedia.org/wiki/Wikipedia:Good_article_criteria}}, we evaluate the article from the aspects of (1) \textit{Interest Level}, (2) \textit{Coherence and Organization}, (3) \textit{Relevance and Focus}, (4) \textit{Coverage}, and (5) \textit{Verifiability}. 
For aspects (1)-(4), we use Prometheus~\citep{kim2023prometheus}, a 13B evaluator LLM to score the article based on a 5-point rubric collaboratively developed with two experienced Wikipedia editors (see Appendix~\ref{appendix:rubric}). For verifiability, we calculate the \textit{citation recall} and \textit{citation precision} based on the definition in~\citet{gao-etal-2023-enabling}. We use Mistral 7B-Instruct~\citep{jiang2023mistral} to examine whether the cited passages entail the generated sentence.

\begin{table*}[h]
\centering
\resizebox{\textwidth}{!}{%
\begin{tabular}{l!{\vrule width \lightrulewidth}ccc!{\vrule width \lightrulewidth}cccc} 
\toprule
                         & \multicolumn{3}{c!{\vrule width \lightrulewidth}}{\textbf{Comparsion with Human-written Articles}} & \multicolumn{4}{c}{\textbf{Rubric Grading}}                     \\
                         & ROUGE-1        & ROUGE-L        & Entity Recall                                                    & Interest Level & Organization  & Relevance     & Coverage       \\ 
\midrule
Direct Gen               & 25.62          & 12.63          & 5.08                                                             & 2.87           & 4.60          & 3.10          & 4.16           \\
RAG     & 28.52          & 13.18          & 7.57                                                             & 3.14           & 4.22          & 3.05          & 4.08           \\
oRAG & 44.26          & 16.51          & 12.57                                                            & 3.90           & 4.79          & 4.09          & 4.70           \\ 
\midrule
\textbf{STORM}           & \textbf{45.82} & \textbf{16.70} & \textbf{14.10}\dag                                                   & \textbf{3.99}\dag  & 4.82 & \textbf{4.45}\dag & \textbf{4.88}\dag  \\
~ w/o Outline Stage      & 26.77          & 12.77          & 7.39                                                             & 3.33           & \textbf{4.87}          & 3.35          & 4.37           \\
\bottomrule
\end{tabular}
}
\caption{Results of automatic article quality evaluation. \dag~ denotes significant differences ($p<0.05$) from a paired $t$-test between \system and the best baseline, \ie, oRAG.
The rubric grading uses a 1-5 scale. 
}
\label{table:article_quality}
\vspace{-0.3em}
\end{table*}

\begin{table}[h]
\centering
\resizebox{\columnwidth}{!}{%
\begin{tabular}{llcc} 
\toprule
                         &                    & \begin{tabular}[c]{@{}c@{}}\textbf{Heading}\\\textbf{Soft Recall}\end{tabular} & \begin{tabular}[c]{@{}c@{}}\textbf{Heading}\\\textbf{Entity Recall}\end{tabular}  \\ 
\midrule

\multirow{5}{*}{GPT-3.5} & Direct Gen         & 80.23                                                                          & 32.39                                                                             \\
                         & RAG/oRAG                & 73.59                                                                          & 33.85                                                                             \\
                         & ~ RAG-expand  & 74.40                                                                          & 33.85                                                                             \\
\cmidrule{2-4}
                         & \textbf{STORM}     & \textbf{86.26}\dag                                                                 & \textbf{40.52}\dag                                                                    \\
                         & ~ w/o Perspective  & 84.49                                                                          & 40.12                                                                             \\
                         & ~ w/o Conversation & 77.97                                                                          & 31.98                                                                             \\

\midrule

\multirow{5}{*}{GPT-4} & Direct Gen         & 87.66                                                                          & 34.78                                                                             \\
                         & RAG/oRAG                & 89.55                                                                          & 42.38                                                                             \\ 
                         & ~ RAG-expand  & 91.36                                                                          & 43.53                                                                             \\
\cmidrule{2-4}
                         & \textbf{STORM}     & \textbf{92.73}\dag                                                                 & \textbf{45.91}                                                                    \\
                         & ~ w/o Perspective  & 92.39                                                                          & 42.70                                                                             \\
                         & ~ w/o Conversation & 88.75                                                                          & 39.30                                                                             \\ 
\bottomrule
\end{tabular}
}
\caption{Results of outline quality evaluation (\%). \dag~ denotes significant differences ($p<0.05$) from a paired $t$-test between \system and baselines.}
\label{table:outline_quality}
\vspace{-0.7em}
\end{table}

\subsection{Baselines}
As prior works use different setups and do not use LLMs, they are hard to compare directly. Instead, we use the following three LLM-based baselines.

\begin{enumerate}
\item {\textit{Direct Gen}, a baseline that directly prompts the LLM to generate an outline, which is then used to generate the full-length article.}
\item {\textit{RAG}, a retrieval-augmented generation baseline that searches with the topic and uses the searched results together with the topic $t$ to generate an outline or the entire article.} 
\item {\textit{Outline-driven RAG (oRAG)}, which is identical to \textit{RAG} in outline creation, but further searches additional information with section titles to generate the article section by section.}
\end{enumerate}

\subsection{\system Implementation}
We build \system with zero-shot prompting using the DSPy framework~\citep{khattab2023dspy}. Appendix~\ref{appendix:pseudo_code} includes the pseudo code and corresponding prompts. The hyperparameters $N$ and $M$ in \system are both set as 5. We use the chat model \texttt{gpt-3.5-turbo} for question asking and use \texttt{gpt-3.5-turbo-instruct} for other parts of \system. We also experiment with using \texttt{gpt-4} for drafting and refining the outline (\reffig{Fig.method}~\Circled{7}-\Circled{8}). For reported results, the simulated topic expert in \system is grounded on the You.com search API\footnote{\url{https://documentation.you.com/api-reference/search}}, although the proposed pipeline is compatible with other search engines. The ground truth Wikipedia article is excluded from the search results.

For final article generation, we only report the results using \texttt{gpt-4} as \texttt{gpt-3.5} is not faithful to sources when generating text with citations~\citep{gao-etal-2023-enabling}. We set temperature as 1.0 and top\_p as 0.9 for all experiments.

\section{Results and Analysis}
\label{sec:results}

\subsection{Main Results}
\label{sec:main_result}

We use outline coverage as a proxy to assess the pre-writing stage (see~\refsec{sec:task_outline}).~\reftab{table:outline_quality} shows the heading soft recall and entity recall. %
Outlines directly generated by LLMs (\textit{Direct Gen}) already demonstrate high heading soft recall, indicating LLMs' ability to grasp high-level aspects of a topic through their rich parametric knowledge. However, \system, by asking effective questions to research the topic, can create higher recall outlines that cover more topic-specific aspects. Notably, although \textit{RAG} leverages additional information, %
presenting unorganized information in the context window makes outline generation more challenging for the weaker model, \ie, GPT-3.5, leading to worse performance. To test the limit of the \textit{RAG} baseline, we further expand the retrieved sources by starting with the outline produced by \textit{RAG}, using its section titles as search queries to collect more sources, and inputting the newly collected sources together with the initial outline to LLM to generate a polished outline. This modified approach is referred to as ``\textit{RAG-expand}'' in~\reftab{table:outline_quality}. The experiment results indicate that even though having an additional round of search and refinement can improve the outline produced by \textit{RAG}, our proposed \system still surpasses its performance.

We further evaluate the full-length article quality. %
As shown in~\reftab{table:article_quality}, \textit{oRAG} significantly outperforms \textit{RAG}, highlighting the effectiveness of using outlines for structuring full-length article generation. Despite this method's advantages in leveraging retrieval and outlining, our approach still outperforms it. %
The effective question asking mechanism %
enhances the articles with greater entity recall. The evaluator LLM also rates these articles with significantly higher scores in the aspects of ``Interest Level'',  ``Relevance and Focus'', and ``Coverage''. Nonetheless, we acknowledge the possibility of the evaluator LLM overrating machine-generated text. Our careful human evaluation (\refsec{sec:human_eval}) reveals that \system still has much room for improvement.

\begin{table}
\centering
\resizebox{0.8\columnwidth}{!}{%
\begin{tabular}{lcc} 
\toprule
                       & \textbf{Citation Recall} & \textbf{Citation Precision}  \\ 
\midrule
\system & 84.83           & 85.18               \\
\bottomrule
\end{tabular}
}
\caption{Citation quality judged by Mistral 7B-Instruct.}
\label{table:citation_quality}
\vspace{-0.5em}
\end{table}

Although this work primarily focuses on the pre-writing stage and does not optimize generating text with citations, we still examine the citation quality of articles produced by our approach. As reported in~\reftab{table:citation_quality}, Mistral 7B-Instruct judges 84.83\% of the sentences are supported by their citations. Appendix~\ref{appendix:citation_quality_discuss} investigates the unsupported sentences and reveals that the primary issues stem from drawing improper inferences and inaccurate paraphrasing, rather than hallucinating non-existent contents.

\subsection{Ablation Studies}
\begin{table}
\centering
\resizebox{\columnwidth}{!}{%
\begin{tabular}{lccc} 
\toprule
                      & \system & w/o Perspective & w/o Conversation  \\ 
\midrule
$|\mathcal{R}|$ & \textbf{99.83} & 54.36           & 39.56             \\
\bottomrule
\end{tabular}
}
\caption{Average number of unique references ($|\mathcal{R}|$) collected using different methods.}
\label{table:ablation_d}
\end{table}

As introduced in~\refsec{sec:method}, \system prompts LLMs to ask effective questions by discovering specific perspectives and simulating multi-turn conversations. We conduct the ablation study on outline creation by comparing \system with two variants: (1) ``\system w/o Perspective'', which omits perspective in the question generation prompt; (2) ``\system w/o Conversation'', which prompts LLMs to generate a set number of questions altogether. To ensure a fair comparison, we control an equal total number of generated questions across all variants. \reftab{table:outline_quality} shows the ablation results and full \system pipeline produces outlines with the highest recall. Also, ``\system w/o Conversation'' gives much worse results, indicating reading relevant information is crucial to generating effective questions. We further examine how many unique sources are collected in $\mathcal{R}$ via different variants. As shown in~\reftab{table:ablation_d}, the full pipeline discovers more different sources and the trend is in accord with the automatic metrics for outline quality.

We also verify whether having an outline stage is necessary with \system. In~\reftab{table:article_quality}, ``\system w/o Outline Stage'' denotes the results of generating the entire article given the topic and the simulated conversations. %
Removing the outline stage significantly deteriorates the performance across all metrics.

\section{Human Evaluation}
\label{sec:human_eval}

\begin{table}
\centering
\resizebox{\columnwidth}{!}{%
\begin{tabular}{lccccc} 
\toprule
                & \multicolumn{2}{c}{oRAG}    & \multicolumn{2}{c}{STORM}       & \multirow{2}{*}{$p$-value}  \\
                & Avg.          & $\geq 4$ Rates & Av.g.         & $\geq 4$ Rates      &                           \\ 
\midrule
Interest Level  & 3.63          & 57.5\%     & \textbf{4.03} & \textbf{70.0\%}          & 0.077                     \\
Organization    & 3.25          & 45.0\%     & \textbf{4.00} & \textbf{70.0\%}          & 0.005                     \\
Relevance       & 3.93          & 62.5\%     & \textbf{4.15} & \textbf{65.0\%}          & 0.347                     \\
Coverage        & 3.58          & 57.5\%     & \textbf{4.00} & \textbf{67.5\%}          & 0.084                     \\
Verifiability   & \textbf{3.85} & 67.5\%     & 3.80          & 67.5\%          & 0.843                     \\ 
\midrule
\#Preferred & \multicolumn{2}{c}{14}     & \multicolumn{2}{c}{\textbf{26}} &                           \\
\bottomrule
\end{tabular}}
\caption{Human evaluation results on 20 pairs of articles generated by \system and \textit{oRAG}. Each pair of articles is evaluated by two Wikipedia editors.
The ratings are given on a scale between 1 and 7, with values $\geq4$ indicating good quality (see~\reftab{table:human_eval_rubric}). We conduct paired $t$-test and report the $p$-value.}
\label{table:human_eval}
\vspace{-0.7em}
\end{table}

To better understand the strengths and weaknesses of \system, we conduct human evaluation by collaborating with 10 experienced Wikipedia editors who have made at least 500 edits on Wikipedia and have more than 1 year of experience. We randomly sample 20 topics from our dataset and evaluate the articles generated by our method and \textit{oRAG}, the best baseline according to the automatic evaluation. Each pair of articles is assigned to 2 editors.

We request editors to judge each article from the same five aspects defined in~\refsec{sec:automatic_metrics}, but using a 1 to 7 scale for more fine-grained evaluation. While our automatic evaluation uses citation quality as a proxy to evaluate \textit{Verifiability}, we stick to the Wikipedia standard of ``verifiable with no original research'' in human evaluation. 
Besides rating the articles, editors are asked to provide open-ended feedback and pairwise preference. After the evaluation finishes, they are further requested to compare an article produced by our method, which they have just reviewed, with its human-written counterpart, and report their perceived usefulness of \system using a 1-5 Likert scale. More human evaluation details are included in Appendix~\ref{appendix:human_eval}. \reftab{table:human_eval} presents the rating and pairwise comparison results.\footnote{For the 1-7 scale rating results on each criterion, we calculate the Krippendorff's Alpha to measure the inter annotator agreement (IAA), and the results are as follows: \textit{Interest Level} (0.349), \textit{Organization} (0.221), \textit{Relevance} (0.256), \textit{Coverage} (0.346), \textit{Verifiability} (0.388).}

\noindent
\textbf{Articles produced by \system exhibit greater breadth and depth than oRAG outputs.} In accord with the finding in %
\refsec{sec:main_result}, editors judge articles produced by \system as more interesting, organized, and having broader coverage compared to %
\textit{oRAG} outputs. Specifically, 25\% more articles produced by \system are considered organized (\textit{Organization} rating $\geq 4$), and 10\% more are deemed to have good coverage (\textit{Coverage} rating $\geq 4$). Even in comparison with human-written articles, one editor praises our result as providing ``a bit more background information'' and another notes that ``I found that the AI articles had more depth compared to the Wikipedia articles''. \system also outperforms the best baseline in pairwise comparison.

\noindent
\textbf{More information in $|\mathcal{R}|$ poses challenges beyond factual hallucination.} We examine 14 pairwise comparison responses where editors prefer oRAG outputs over \system. Excluding 3 cases where pairwise preferences do not align with their ratings, editors assign lower \textit{Verifiability} scores to articles from our approach in over 50\% of the cases. Through analyzing the articles and editors' free-form feedback, 
we discover that \textit{low Verifiability scores stem from red herring fallacy or overspeculation issues}. These arise when the generated articles introduce unverifiable connections between different pieces of information in $|\mathcal{R}|$ or between the information and the topic (examples included in~\reftab{table:comment_summary}). Compared to the widely discussed factual hallucination~\citep{shuster-etal-2021-retrieval-augmentation,huang2023survey}, addressing such verifiability issues is more nuanced, surpassing basic fact-checking~\citep{min-etal-2023-factscore}. %

\begin{figure}[t]
    \centering 
    \resizebox{\columnwidth}{!}{%
    \includegraphics{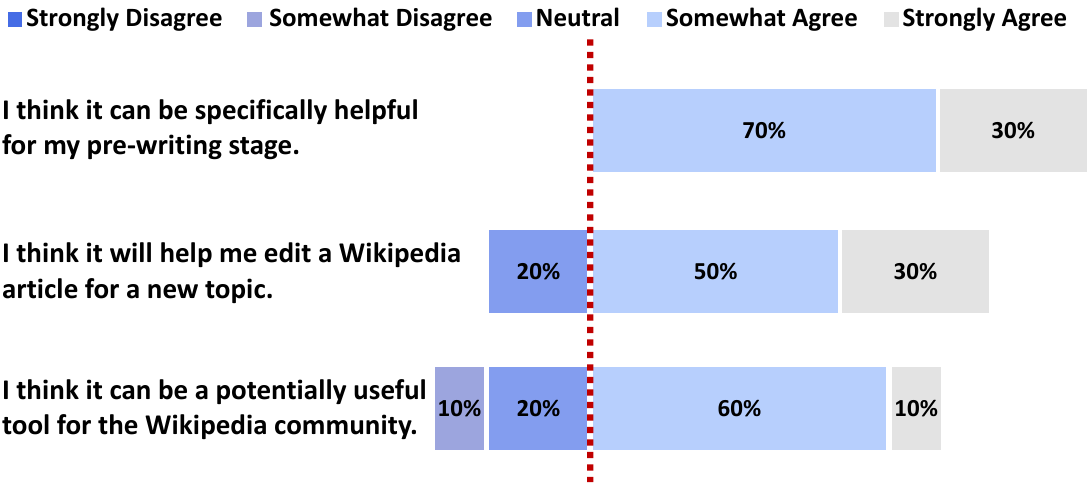}
    }
    \caption{Survey results of the perceived usefulness of \system ($n=10$).
    }
    \label{Fig.survey}
\vspace{-0.5em}
\end{figure}

\noindent
\textbf{Generated articles trail behind well-revised human works.}
While \system outperforms the oRAG baseline, editors comment that the generated articles are \textit{less informative than actual Wikipedia pages}. Another major issue identified is \textit{the transfer of bias and tone from Internet sources to the generated article}, with 7 out of 10 editors mentioning that the STORM-generated articles sound ``emotional'' or ``unneutral''. More analysis is discussed in Appendix~\ref{appendix:error_analysis}. This feedback suggests that reducing the retrieval bias in the pre-writing stage is a worthwhile direction for future work.

\noindent
\textbf{Generated articles %
are a good starting point.} As shown in~\reffig{Fig.survey}, editors are unanimous in agreeing that \system can aid them in their pre-writing stage.
It is gratifying to know that the tool is helpful to experienced editors.
80\% of the editors think that \system can help them edit a Wikipedia article for a new topic. More reservation is expressed to the usefulness of \system for the Wikipedia community at large; nonetheless, 70\% of the editors think it is useful, with only 10\% disagreeing.

\section{Related Works}
\label{sec:related_work}
\noindent
\textbf{Retrieval-Augmented Generation (RAG)}\quad
Augmenting language models (LMs) with retrieval at inference time is a typical way to leverage external knowledge stores%
~\citep{ram2023ralm,JMLR:v24:23-0037}. While some works use retrieval to construct demonstrations for in-context learning~\citep{li-etal-2023-unified,liu-etal-2022-makes,agrawal-etal-2023-context,poesia2022synchromesh,shi-etal-2022-nearest,khattab2022demonstrate}, another line of works uses retrieval to provide additional information for LMs to ground on.~\citet{lewis2020retrieval} study RAG on knowledge-intensive NLP tasks and find it improves diversity and factuality. \citet{semnani-etal-2023-wikichat} designs a RAG-based chatbot grounded on English Wikipedia to stop LLM-based chatbots from hallucination. Besides, RAG can be used to generate text with citations~\citep{menick2022teaching,gao-etal-2023-enabling} and build attributed question answering systems~\citep{bohnet2023attributed}. %
While RAG is widely studied in question answering, how to use it for long-form article generation is less investigated.%

As a general framework, RAG is flexible in both the retrieval source and time. The retrieval sources can vary from domain databases~\citep{zakka2023almanac}, code documentation~\citep{zhou2023docprompting}, to the whole Internet~\citep{nakano2022webgpt,komeili-etal-2022-internet}. Regarding the time, %
besides a one-time retrieval before generation, the system can be designed to self-decide when %
to retrieve across the course of the generation~\citep{jiang-etal-2023-active, parisi2022talm,shuster-etal-2022-language,yao2023react}. %

\noindent
\textbf{Automatic Expository Writing}\quad
Different from other types of long-form generation~\citep{yang-etal-2022-re3, feng2018topic}, %
automatic expository writing requires grounding on external documents and leveraging the interplay between reading and writing. \citet{balepur-etal-2023-expository} propose the Imitate-Retrieve-Paraphrase framework for expository writing at the paragraph level to address the challenges in synthesizing information from multiple sources. Beyond summarizing sources, 
\citet{shen2023summarization} highlight that expository writing requires the author's sensemaking process over source documents and good outline planning. We tackle these challenges by focusing on the pre-writing stage.

\noindent
\textbf{Question Asking in NLP}\quad
Question asking capabilities in NLP systems have expanded across several fronts, including generating clarification questions to understand user intents~\citep{aliannejadi2019asking, rahmani-etal-2023-survey}, and breaking large questions into smaller ones to improve compositional reasoning~\citep{press-etal-2023-measuring}. While humans usually ask questions to learn new knowledge~\citep{tawfik2020role,booth2003craft}, how to optimize question informativeness and specificity in information-seeking conversations remains less explored. The closest work is~\citet{qi-etal-2020-stay} which defines the question informativeness using the unigram precision function and uses reinforcement learning to increase the question informativeness.

\section{Conclusion}
\label{sec:conclusion}
We propose \system, an LLM-based writing system that automates the pre-writing stage for creating Wikipedia-like articles from scratch. We curate the FreshWiki dataset and establish evaluation criteria to study the generation of grounded long-form articles. Experimental results demonstrate that the question asking mechanism in \system improves both the outline and article quality. 
With the improved breadth and depth, \system helps surface new challenges for grounded writing systems through expert evaluation. The experienced Wikipedia editors in our study unanimously agree that \system is helpful for their pre-writing stage.

\section*{Limitations}
In this work, we explore generating Wikipedia-like articles from scratch as a way to push the frontier of automatic expository writing and long-form article generation. While our approach significantly outperforms baseline methods in both automatic and human evaluations, the quality of machine-written articles still lags behind well-revised human-authored articles, specifically in aspects of neutrality and verifiability. Although \system discovers different perspectives in researching the given topic, the collected information may still be biased towards dominant sources on the Internet and may contain promotional content. Moreover, the verifiability issues identified in this work go beyond factual hallucination, which highlights new challenges to grounded writing systems.

Another limitation of this work is that although we focus on the task of generating Wikipedia-like articles from scratch, our task setup is still simplified to only consider the generation of free-form text. Human-authored high-quality Wikipedia articles usually contain structured data and multi-modal information. We leave the exploration of generating multi-modal grounded articles for future work.

\section*{Acknowledgements}
We thank You.com for generously providing the search API that supported our experiments. We also thank Sina J. Semnani, Shicheng Liu, Eric Zelikman for providing helpful feedback and the ACL ARR reviewers for their valuable comments. This work is supported in part by the Verdant Foundation and Microsoft Azure AI credits. Yijia Shao is supported by a Stanford School of Engineering Fellowship.

\section*{Ethics Statement}
Different from the creative generation, grounded article generation may impact how people learn about topics or consume source information. All the studies and the evaluation in this work are designed to prevent the dissemination of misinformation by not publishing generated content online and implementing strict accuracy checks. We avoid any disruption to Wikipedia or related communities, as our system does not interact with live pages. Also, although we try to generate grounded articles, we believe there is no privacy issue related to this work as we only use information publicly available on the Internet.

The primary risk of our work is that the Wikipedia articles written by our system are grounded on information on the Internet which contains some biased or discriminative content on its own. Currently, our system relies on the search engine to retrieve information but does not include any post-processing module. We believe improving the retrieval module to have good coverage of different viewpoints and adding a content sifting module to the current system will be a critical next step to achieve better neutrality and balance in the generated articles. %

Another limitation we see from an ethical point of view is that we only consider writing English Wikipedia articles in this work. Extending the current system to a multilingual setup is a meaningful direction for future work as more topics do not have Wikipedia pages in non-English languages.

\bibliography{anthology,custom}

\clearpage
\appendix
\label{sec:appendix}
\section{Dataset Details}
\label{appendix:dataset}

As discussed in~\refsec{sec:task_dataset}, we curate the FreshWiki dataset by collecting \textit{recent} and \textit{high-quality} English Wikipedia articles. We select the most-edited pages over a specific period rather than using creation dates as a cutoff because most of Wikipedia articles are ``stubs'' or are of low quality when they were created. For quality, we consider articles predicted to be of B-class quality or above. According to Wikipedia statistics\footnote{\url{https://en.wikipedia.org/wiki/Wikipedia:Content_assessment}}, only around 3\% of existing Wikipedia pages meet this quality standard. As LLMs can generate reasonably good outputs, we think it is important to use high-quality human-written articles as references for further research.

\begin{table}
\centering
\resizebox{0.85\columnwidth}{!}{%
\begin{tabular}{lc} 
\toprule
Average Numer of Sections            & 8.4     \\
Average Number of All-level Headings & 15.8    \\ 
\midrule
Average Length of a Section          & 327.8   \\
Average Length of Total Article      & 2159.1  \\ 
\midrule
Average Number of References         & 90.1    \\
\bottomrule
\end{tabular}
}
\caption{Statistics of the dataset used in our experiments.}
\label{table:data_statistics}
\end{table}

\begin{figure}[t]
    \centering 
    \resizebox{\columnwidth}{!}{%
    \includegraphics{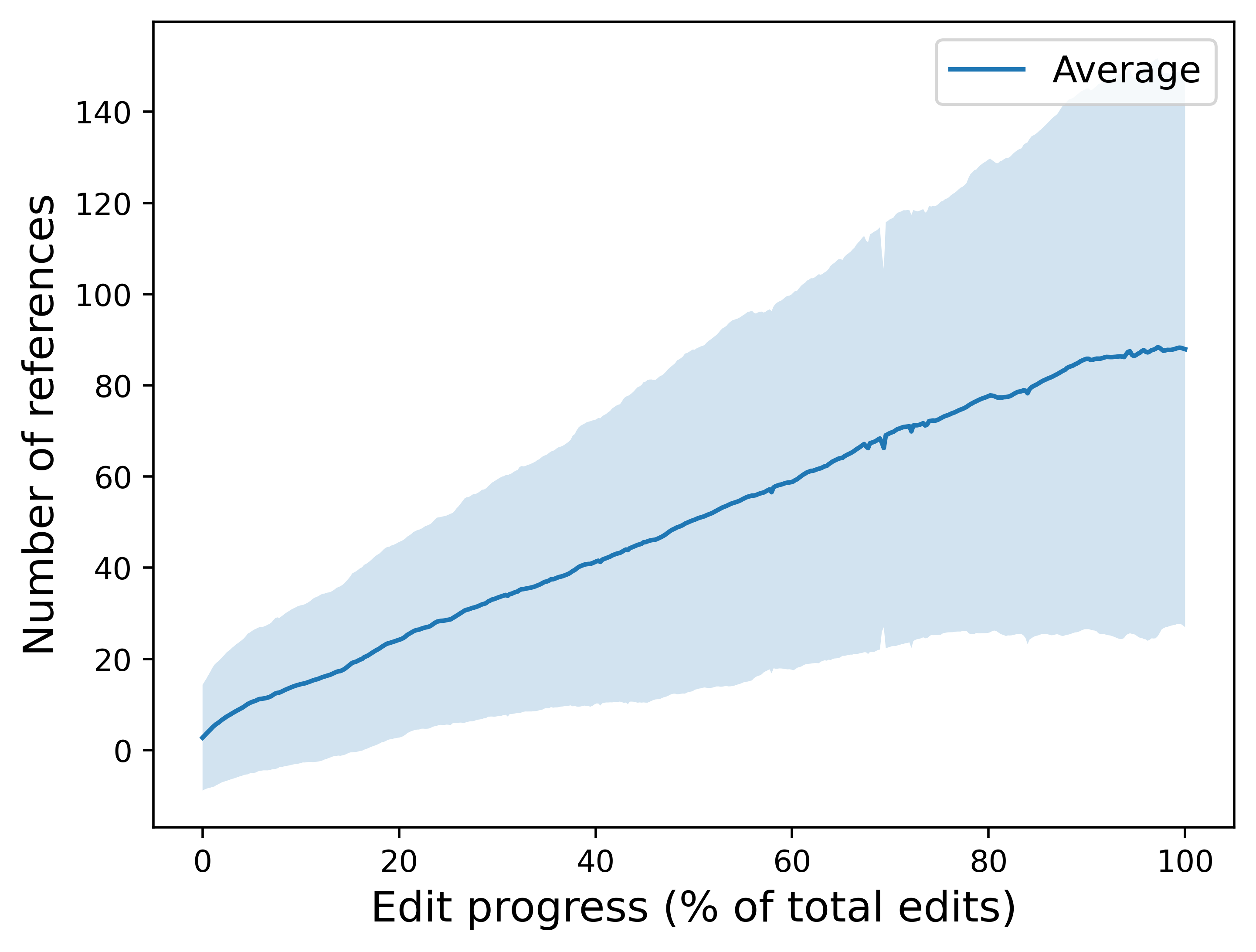}
    }
    \caption{Evolution of reference count in the Wikipedia article editing process.
    }
    \label{Fig.ref_trend}
\end{figure}

\begin{figure}[t]
    \centering 
    \resizebox{\columnwidth}{!}{%
    \includegraphics{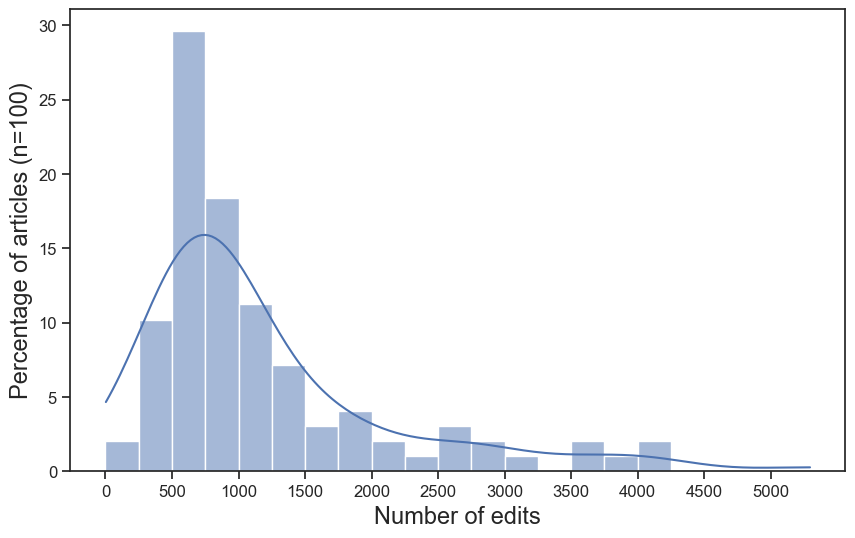}
    }
    \caption{Distribution of edit counts for Wikipedia articles in our experiments ($n=100$). 
    }
    \label{Fig.edit_distribution}
\end{figure}

For experiments in this work, we randomly select 100 samples with human-written articles under 3000 words to have a meaningful comparison. \reftab{table:data_statistics} gives the data statistics. Notably, human-authored articles have a large number of references but they require numerous edits to achieve this. \reffig{Fig.ref_trend} illustrates the evolution of the reference count in the article edit process and \reffig{Fig.edit_distribution} gives the distribution of edit counts for human-authored articles used in our experiments.

\section{Pseudo Code of \system}
\label{appendix:pseudo_code}
In~\refsec{sec:method}, we introduce \system, a framework that automates the pre-writing stage by discovering different perspectives, simulating information-seeking conversations, and creating a comprehensive outline. Algorithm~\ref{code:pipeline} displays the skeleton of \system.

We implement \system with zero-shot prompting using the DSPy framework~\citep{khattab2023dspy}. Listing~\ref{lst:prompt1} and~\ref{lst:prompt2} show the prompts used in our implementation. We highlight that \system offers a general framework designed to assist the creation of grounded, long-form articles, without depending extensively on prompt engineering for a single domain.

\begin{algorithm}
    \SetKwFunction{findRelatedTopics}{gen\_related\_topics}
    \SetKwFunction{getWikiArticle}{get\_wiki\_article}
    \SetKwFunction{extractTOC}{extract\_toc}
    \SetKwFunction{identifyPerspectives}{gen\_perspectives}
    \SetKwFunction{askQuestion}{gen\_qn}
    \SetKwFunction{questionToQuery}{gen\_queries}
    \SetKwFunction{search}{search\_and\_sift}
    \SetKwFunction{answer}{gen\_ans}
    \SetKwFunction{genOutline}{direct\_gen\_outline}
    \SetKwFunction{refineOutline}{refine\_outline}
    \SetKwInOut{KwIn}{Input}
    \SetKwInOut{KwOut}{Output}

    \KwIn{Topic $t$, maximum perspective $N$, maximum conversation round $M$}
    \KwOut{Outline $\mathcal{O}$, references $\mathcal{R}$}

    P0 = "basic fact writer ..." \tcp{Constant.}

    $\mathcal{R} \leftarrow [\ ]$

    \tcp{Discover perspectives $\mathcal{P}$.}
    
    related\_topics $\leftarrow$ $\findRelatedTopics$($t$)

    tocs $\leftarrow [\ ]$ 

    \ForEach{\upshape{related\_t} in related\_topics}{
        article $\leftarrow$ $\getWikiArticle$(related\_t)
        \If{\upshape{article}}{
            tocs.append($\extractTOC$(article))
        }
    }

    $\mathcal{P} \leftarrow$ $\identifyPerspectives$($t$, tocs)

    $\mathcal{P} \leftarrow$ [P0] + $\mathcal{P}$[:$N$]

    \tcp{Simulate conversations.}
    convos $\leftarrow$ $[\ ]$
    
    \ForEach{$p$ in $\mathcal{P}$}{
        convo\_history $\leftarrow$ $[\ ]$
        
        \For{$i$ = $1$ \upshape{to} $M$}{
            \tcp{Question asking.}
            
            $q$ $\leftarrow$ $\askQuestion$($t$, $p$, dlg\_history)
            
            convo\_history.append($q$)

            \tcp{Question answering.}

            queries $\leftarrow$ $\questionToQuery$($t$, q)

            sources $\leftarrow$ $\search$(queries)

            $a$ $\leftarrow$ $\answer$($t$, q, sources)

            convo\_history.append($a$)

            $\mathcal{R}$.append(sources)
        }

        convos.append(convo\_history)

    }

    \tcp{Create the outline.}

    $\mathcal{O}_\text{D}$ $\leftarrow$ $\genOutline$($t$)

    $\mathcal{O}$ $\leftarrow$ $\refineOutline$($t$, $\mathcal{O}_\text{D}$, convos)
    
    \KwRet{$\mathcal{O}$, $\mathcal{R}$}

    \caption{\system}
    \label{code:pipeline}
\end{algorithm}

\begin{figure*}
    \begin{lstlisting}[language=Python]
class GenRelatedTopicsPrompt(dspy.Signature):
    """
    I'm writing a Wikipedia page for a topic mentioned below. Please identify and recommend some Wikipedia pages on closely related subjects. I'm looking for examples that provide insights into interesting aspects commonly associated with this topic, or examples that help me understand the typical content and structure included in Wikipedia pages for similar topics.
    Please list the urls in separate lines.
    """

    topic = dspy.InputField(prefix="Topic of interest:", format=str)
    related_topics = dspy.OutputField()

class GenPerspectivesPrompt(dspy.Signature):
    """
    You need to select a group of Wikipedia editors who will work together to create a comprehensive article on the topic. Each of them represents a different perspective, role, or affiliation related to this topic. You can use other Wikipedia pages of related topics for inspiration. For each editor, add description of what they will focus on.
    Give your answer in the following format: 1. short summary of editor 1: description\n2. short summary of editor 2: description\n...
    """

    topic = dspy.InputField(prefix='Topic of interest:', format=str)
    examples = dspy.InputField(prefix='Wiki page outlines of related topics for inspiration:\n', format=str)
    perspectives = dspy.OutputField()

class GenQnPrompt(dspy.Signature):
    """
    You are an experienced Wikipedia writer and want to edit a specific page. Besides your identity as a Wikipedia writer, you have a specific focus when researching the topic.
    Now, you are chatting with an expert to get information. Ask good questions to get more useful information.
    When you have no more question to ask, say "Thank you so much for your help!" to end the conversation.
    Please only ask one question at a time and don't ask what you have asked before. Your questions should be related to the topic you want to write.
    """

    topic = dspy.InputField(prefix='Topic you want to write: ', format=str)
    persona = dspy.InputField(prefix='Your specific perspective: ', format=str)
    conv = dspy.InputField(prefix='Conversation history:\n', format=str)
    question = dspy.OutputField()

class GenQueriesPrompt(dspy.Signature):
    """
    You want to answer the question using Google search. What do you type in the search box?
    Write the queries you will use in the following format:- query 1\n- query 2\n...
    """

    topic = dspy.InputField(prefix='Topic you are discussing about: ', format=str)
    question = dspy.InputField(prefix='Question you want to answer: ', format=str)
    queries = dspy.OutputField()

    \end{lstlisting}
    \captionof{lstlisting}{Prompts used in \system, corresponding to Line 4, 11, 19, 22 in Algorithm~\ref{code:pipeline}.}
    \label{lst:prompt1}
\end{figure*}

\begin{figure*}
    \begin{lstlisting}[language=Python]
class GenAnswerPrompt(dspy.Signature):
    """
    You are an expert who can use information effectively. You are chatting with a Wikipedia writer who wants to write a Wikipedia page on topic you know. You have gathered the related information and will now use the information to form a response.
    Make your response as informative as possible and make sure every sentence is supported by the gathered information.
    """

    topic = dspy.InputField(prefix='Topic you are discussing about:', format=str)
    conv = dspy.InputField(prefix='Question:\n', format=str)
    info = dspy.InputField(
        prefix='Gathered information:\n', format=str)
    answer = dspy.OutputField(prefix='Now give your response:\n')


class DirectGenOutlinePrompt(dspy.Signature):
    """
    Write an outline for a Wikipedia page.
    Here is the format of your writing:
        1. Use "#" Title" to indicate section title, "##" Title" to indicate subsection title, "###" Title" to indicate subsubsection title, and so on.
        2. Do not include other information.
    """

    topic = dspy.InputField(prefix="Topic you want to write: ", format=str)
    outline = dspy.OutputField(prefix="Write the Wikipedia page outline:\n")


class RefineOutlinePrompt(dspy.Signature):
    """
    Improve an outline for a Wikipedia page. You already have a draft outline that covers the general information. Now you want to improve it based on the information learned from an information-seeking conversation to make it more comprehensive.
    Here is the format of your writing:
        1. Use "#" Title" to indicate section title, "##" Title" to indicate subsection title, "###" Title" to indicate subsubsection title, and so on.
        2. Do not include other information.
    """

    topic = dspy.InputField(prefix="Topic you want to write: ", format=str)
    conv = dspy.InputField(prefix="Conversation history:\n", format=str)
    old_outline = dspy.OutputField(prefix="Current outline:\n", format=str)
    outline = dspy.OutputField(
        prefix='Write the Wikipedia page outline:\n')

    \end{lstlisting}
    \captionof{lstlisting}{Prompts used in \system (continue), corresponding to Line 24, 31, 32 in Algorithm~\ref{code:pipeline}.}
    \label{lst:prompt2}
\end{figure*}

\section{Automatic Evaluation Details}
\subsection{Soft Heading Recall}
\label{appendix:soft_heading_recall}
We calculate the soft heading recall between the multi-level headings in the generated outline, considered as the prediction $P$, and those in the human-written article, considered as the ground truth $G$. The calculation is based on the soft recall definition in~\citet{franti2023soft}. Given a set $A=\{Ai\}_{i=1}^K$, \textit{soft count} of an item is defined as the inverse of the sum of its similarity to other items in the set:
\begin{equation}
\resizebox{0.8\columnwidth}{!}{$%
\begin{gathered}
    \operatorname{count}\left(A_i\right)=\frac{1}{\sum_{j=1}^K \operatorname{Sim}\left(A_i, A_j\right)} \\
    \operatorname{Sim}\left(A_i, A_j\right) = \operatorname{cos}\left(\operatorname{embed}(A_i), \operatorname{embed}(A_j)\right),
\end{gathered}$%
}
\label{eq:soft_count}
\end{equation}
where $\operatorname{embed}(\cdot)$ in \refequ{eq:soft_count} is parameterized by \texttt{paraphrase-MiniLM-L6-v2} provided in the Sentence-Transformers library\footnote{\url{https://huggingface.co/sentence-transformers/paraphrase-MiniLM-L6-v2}}. The cardinality of $A$ is the sum of the counts of its individual items:
\begin{equation}
    \operatorname{card}(A)=\sum_{i=1}^K \operatorname{count}\left(A_i\right)
\end{equation}

The soft heading recall is calculated as
\begin{equation}
    \text{\textit{soft heading recall}} = \frac{\operatorname{card}(G \cap P)}{\operatorname{card}(G)},
\end{equation}
where the cardinality of intersection is defined via the union as follows:
\begin{equation}
\resizebox{0.75\columnwidth}{!}{$%
\begin{gathered}
    \operatorname{card}(G \cap P)=\\ \operatorname{card}(G)+\operatorname{card}(P)-\operatorname{card}(G \cup P).
\end{gathered}$%
}
\end{equation}

\subsection{LLM Evaluator}
\label{appendix:rubric}

\begin{table*}
\centering
\resizebox{\textwidth}{!}{%
\begin{tabular}{ll} 
\toprule
Criteria Description & \textbf{Interest Level}: How engaging and thought-provoking is the article?                                                                                                               \\
Score 1 Description  & Not engaging at all; no attempt to capture the reader's attention.                                                                                                                        \\
Score 2 Description  & Fairly engaging with a basic narrative but lacking depth.                                                                                                                                 \\
Score 3 Description  & Moderately engaging with several interesting points.                                                                                                 \\
Score 4 Description  & Quite engaging with a well-structured narrative and noteworthy points that frequently capture and retain attention.                                                                       \\
Score 5 Description  & Exceptionally engaging throughout, with a compelling narrative that consistently stimulates interest.                                                                                     \\ 
\midrule
Criteria Description & \textbf{Coherence and Organization}: Is the article well-organized and logically structured?                                                                                              \\
Score 1 Description  & Disorganized; lacks logical structure and coherence.                                                                                                                                      \\
Score 2 Description  & Fairly organized; a basic structure is present but not consistently followed.                                                                                                             \\
Score 3 Description  & Organized; a clear structure is mostly followed with some lapses in coherence.                                                                                                 \\
Score 4 Description  & Good organization; a clear structure with minor lapses in coherence.                                                                                                                      \\
Score 5 Description  & Excellently organized; the article is logically structured with seamless transitions and a clear argument.                                                                                \\ 
\midrule
Criteria Description & \textbf{Relevance and Focus}: Does the article stay on topic and maintain a clear focus?                                                                                                  \\
Score 1 Description  & Off-topic; the content does not align with the headline or core subject.                                                                                                                  \\
Score 2 Description  & Somewhat on topic but with several digressions; the core subject is evident but not consistently adhered to.                                                                              \\
Score 3 Description  & Generally on topic, despite a few unrelated details.                                                                                                                                      \\
Score 4 Description  & Mostly on topic and focused; the narrative has a consistent relevance to the core subject with infrequent digressions.                                                                    \\
Score 5 Description  & \makecell[l]{Exceptionally focused and entirely on topic; the article is tightly centered on the subject, with every piece of information contributing \\to a comprehensive understanding of the topic.}  \\ 
\midrule
Criteria Description & \textbf{Broad Coverage}: Does the article provide an in-depth exploration of the topic and have good coverage?                                                                            \\
Score 1 Description  & Severely lacking; offers little to no coverage of the topic's primary aspects, resulting in a very narrow perspective.                                                                    \\
Score 2 Description  & Partial coverage; includes some of the topic's main aspects but misses others, resulting in an incomplete portrayal.                                                                      \\
Score 3 Description  & Acceptable breadth; covers most main aspects, though it may stray into minor unnecessary details or overlook some relevant points.                                                        \\
Score 4 Description  & Good coverage; achieves broad coverage of the topic, hitting on all major points with minimal extraneous information.                                                                     \\
Score 5 Description  & \makecell[l]{Exemplary in breadth; delivers outstanding coverage, thoroughly detailing all crucial aspects of the topic without including irrelevant \\information.}                                      \\
\bottomrule
\end{tabular}
}
\caption{Scoring rubrics on a 1-5 scale for the evaluator LLM.}
\label{table:rubric}
\end{table*}

We use Prometheus\footnote{\url{https://huggingface.co/kaist-ai/prometheus-13b-v1.0}}~\citep{kim2023prometheus}, a 13B open-source evaluator LLM that can assess long-form text based on customized 1-5 scale rubric, to grade the article from the aspects of \textit{Interest level}, \textit{Coherence and Organization}, \textit{Relevance and Focus}, and \textit{Coverage}. \reftab{table:rubric} gives our grading rubric. While Prometheus is best used with a score 5 reference answer, we find adding the reference will exceed the context length limit of the model. Since \citet{kim2023prometheus} show Prometheus ratings without reference also correlate well with human preferences, we omit the reference and trim the input article to be within 2000 words by iteratively removing contents from the shortest section to ensure the input can fit into the model's context window.

\subsection{More Discussion of the Citation Quality}
\label{appendix:citation_quality_discuss}

\begin{table*}
\centering
\resizebox{\textwidth}{!}{%
\begin{tabular}{llll} 
\toprule
\textbf{Error Type}          & \textbf{Topic}                                                    & \textbf{Unsupported Sentence}                   & \textbf{Source}   \\
\midrule
Improper Inferential Linking & Lahaina, Hawaii                                                   & \makecell[l]{Throughout its history, religion has remained the\\ paramount aspect of Hawaiian life \colorbox{myyellow}{in Lahaina},\\ permeating every daily activity and significant event[5].}                                                       & \makecell[l]{[5] ``Religion, Beliefs \& Spirituality'' \\ \textit{(The source discusses religion as part of Hawaiian life}\\ \textit{but \colorbox{myyellow}{does not mention Lahania}.)}}                                      \\
\midrule[0.5pt]
Inaccurate Paraphrasing      & \makecell[l]{2022 Crimean\\Bridge explosion} & \makecell[l]{\colorbox{myyellow}{Completed in June 2020}, the bridge serves as a\\ major supply route for Russian forces in the region\\ and is significant to Russia's claim over the disputed\\ territory[2][11].} & \makecell[l]{[2] ``Crimean Bridge - Wikipedia'' \\ \textit{(The source says ``The first scheduled passenger train}\\ \textit{crossed the bridge on 25 December 2019, while the}\\ \textit{bridge was \colorbox{myyellow}{opened for freight trains on 30 June 2020}''.)}}\\
\midrule[0.5pt]
Citing Irrelevant Sources    & LK-99                                                             & \makecell[l]{For example, comparisons have been drawn between\\ the performance of LK-9 and the dynamic resolution\\ capabilities of video games such as Battlefield 2042[22].}                                                  & \makecell[l]{[22] ``Battlefield 2042 PC performance guide: The best\\ settings for a high frame rate''\\ \textit{(\colorbox{myyellow}{The source is irrelevant to LK-99.})}}  \\
\bottomrule
\end{tabular}
}
\caption{Examples of different error types of unsupported sentences.}
\label{table:example_unsupported}
\end{table*}

\begin{figure}[h]
    \centering 
    \resizebox{0.95\columnwidth}{!}{%
    \includegraphics{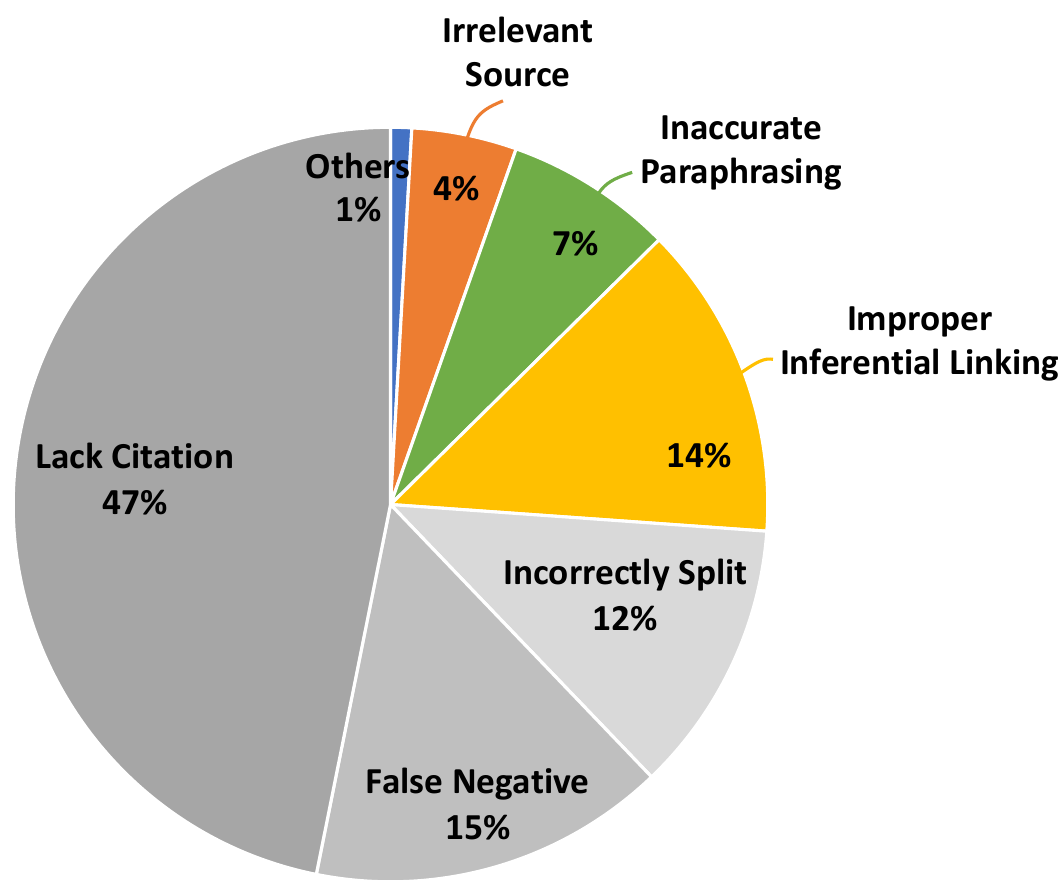}
    }
    \caption{Error analysis of unsupported sentences in 10 sampled articles.
    }
    \label{Fig.citation_error}
\end{figure}

We use Mistral 7B-Instruct\footnote{\url{https://huggingface.co/mistralai/Mistral-7B-Instruct-v0.1}}~\citep{jiang2023mistral} to examine whether the cited passages entail the generated sentence. 
\reftab{table:citation_quality} reports the citation quality of articles produced by our approach, showing that around 15\% sentences in generated articles are unsupported by citations. We further investigate the failure cases by randomly sampling 10 articles and an author manually examines all the unsupported sentences in these articles. Besides sentences that are incorrectly split\footnote{Following~\citet{gao-etal-2023-enabling}, we check citation quality in the sentence level and split articles into sentences using NLTK \texttt{sent\_tokenize}. \texttt{sent\_tokenize} sometimes fails to split sentences correctly when the article contains special words like ``No.12847'', ``Bhatia et al.'', \etc}, lack citations, or are deemed supported by the author's judgment, our analysis identifies three main error categories (examples are given in~\reftab{table:example_unsupported}): \textit{improper inferential linking}, \textit{inaccurate paraphrasing}, and \textit{citing irrelevant sources}. 

We show the error distribution in~\reffig{Fig.citation_error}. Notably, the most common errors stem from the tendency of LLMs to form improper inferential links between different pieces of information presented in the context window. Our analysis of citation quality suggests that, in addition to avoiding hallucinations, future research in grounded text generation should also focus on preventing LLMs from making overly inferential leaps based on the provided information.

\section{Human Evaluation Details}
\label{appendix:human_eval}

We recruited 10 experienced Wikipedia editors to participate in our study by creating a research page on \texttt{Meta-Wiki}\footnote{\url{https://meta.wikimedia.org}} and reaching out to active editors who have recently approved articles for Wikipedia.\footnote{Since evaluating Wikipedia-like articles is time-consuming and requires expertise, we paid each participant 50\$ for our study.} Our participation group includes 3 editors with 1-5 years of experience, 4 with 6-10 years, and 3 with over 15 years of contribution. The study was approved by the Institutional Review Board of our institution and the participants signed the consent form through Qualtrics questionnaires before the study started.

\begin{figure*}[t]
    \centering 
    \resizebox{\textwidth}{!}{%
    \includegraphics{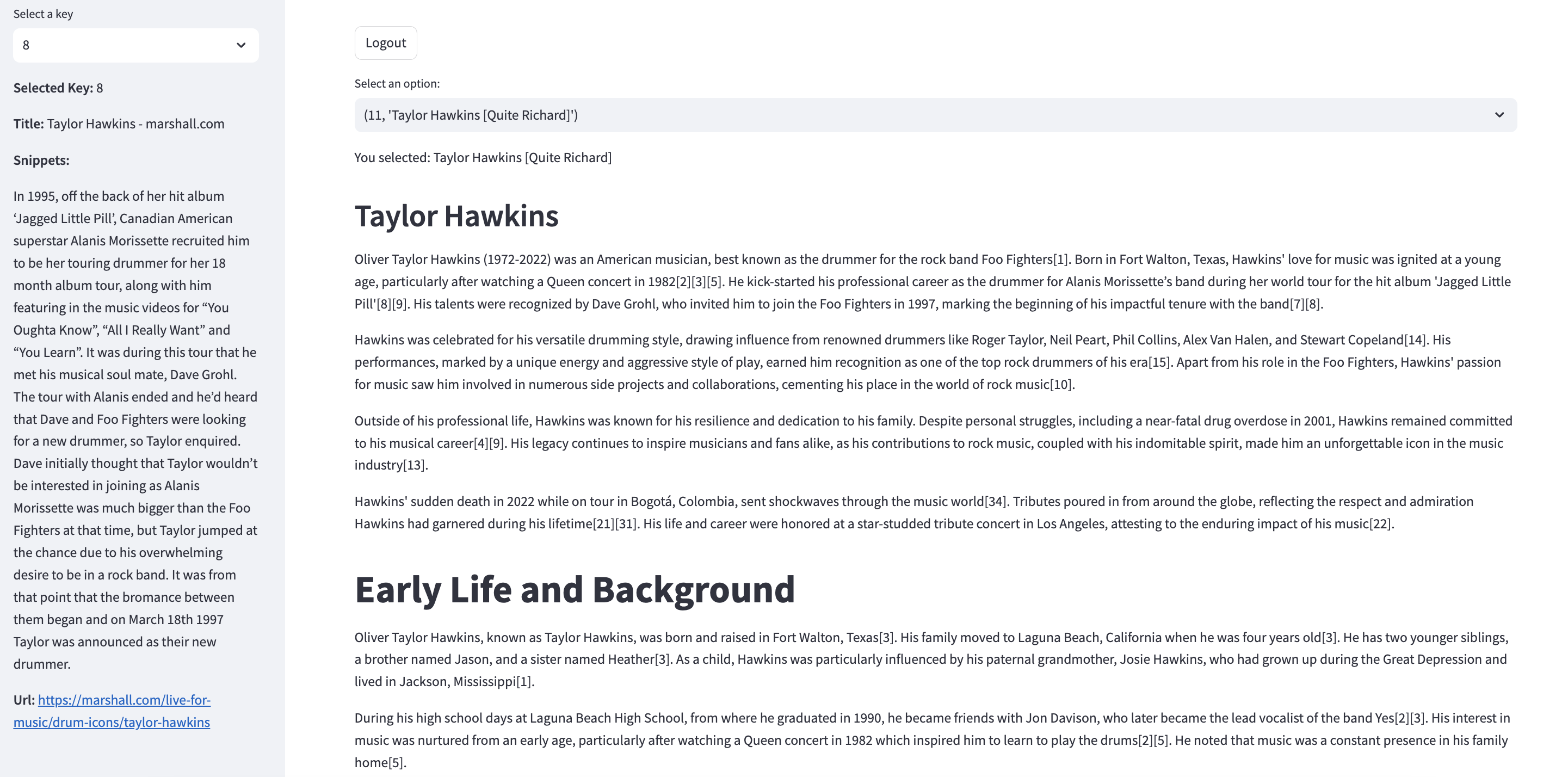}
    }
    \caption{Screenshot of the web application for evaluating the grounded article.
    }
    \label{Fig.ui}
\end{figure*}

To streamline the evaluation of grounded articles, we developed a web application, which features a side-by-side display of the article and its citation snippets, to gather ratings and open-ended feedback for each article. \reffig{Fig.ui} shows the screenshot of our web application and the full article produced by \system is included in~\reftab{table:example}. For human evaluation, we use a 1 to 7 scale for more fine-grained evaluation. The grading rubric is included in~\reftab{table:human_eval_rubric}.

We collected the pairwise preferences and the perceived usefulness of \system via an online questionnaire. Specifically, for the perceived usefulness, we request editors to rate their agreement with statements ``I think it can be specifically helpful for my pre-writing stage (\eg, collecting relevant sources, outlining, drafting).'', ``I think it will help me edit a Wikipedia article for a new topic'', ``I think it can be a potentially useful tool for the Wikipedia community'' on a Likert scale of 1-5, corresponding to \textit{Strongly disagree}, \textit{Somewhat disagree}, \textit{Neither agree nor disagree}, \textit{Somewhat agree}, \textit{Strongly agree}.

\begin{table*}
\centering
\resizebox{\textwidth}{!}{%
\begin{tabular}{l} 
\toprule
\multicolumn{1}{c}{\textbf{Interest Level}}                                                                                                                                                  \\ 
\midrule
1:~Not engaging at all; no attempt to capture the reader's attention.                                                                                                                        \\
2:~Slightly engaging with rare moments that capture attention.                                                                                                                               \\
3:~Fairly engaging with a basic narrative but lacking depth.                                                                                                                                 \\
4:~Moderately engaging with several interesting points.                                                                                                                                      \\
5:~Quite engaging with a well-structured narrative and noteworthy points that frequently capture and retain attention.                                                                       \\
6:~Very engaging with a compelling narrative that captures and mostly retains attention.                                                                                                     \\
7:~Exceptionally engaging throughout, with a compelling narrative that consistently stimulates interest.                                                                                     \\ 
\midrule
\multicolumn{1}{c}{\textbf{Coherence and Organization}}                                                                                                                                      \\ 
\midrule
1:~Disorganized; lacks logical structure and coherence.                                                                                                                                      \\
2:~Poor organization; some structure is evident but very weak.                                                                                                                               \\
3:~Fairly organized; a basic structure is present but not consistently followed.                                                                                                             \\
4:~Organized; a clear structure is mostly followed with some lapses in coherence.                                                                                                            \\
5:~Good organization; a clear structure with minor lapses in coherence.                                                                                                                      \\
6:~Very well-organized; a logical structure with transitions that effectively guide the reader.                                                                                              \\
7:~Excellently organized; the article is logically structured with seamless transitions and a clear argument.                                                                                \\ 
\midrule
\multicolumn{1}{c}{\textbf{Relevance and Focus}}                                                                                                                                             \\ 
\midrule
1:~Off-topic; the content does not align with the headline or core subject.                                                                                                                  \\
2:~Mostly off-topic with some relevant points.                                                                                                                                               \\
3:~Somewhat on topic but with several digressions; the core subject is evident but not consistently adhered to.                                                                              \\
4:~Generally on topic, despite a few unrelated details.                                                                                                                                      \\
5:~Mostly on topic and focused; the narrative has a consistent relevance to the core subject with infrequent digressions.                                                                    \\
6:~Highly relevant with a focused narrative and purpose.                                                                                                                                     \\
\makecell[l]{7:~Exceptionally focused and entirely on topic; the article is tightly centered on the subject, with every piece of information contributing to a\\ comprehensive understanding of the topic.}  \\ 
\midrule
\multicolumn{1}{c}{\textbf{Broad Coverage}}                                                                                                                                                  \\ 
\midrule
1:~Severely lacking; offers little to no coverage of the topic's primary aspects, resulting in a very narrow perspective.                                                                    \\
2:~Minimal coverage; addresses only a small selection of the topic's main aspects, with significant omissions.                                                                               \\
3:~Partial coverage; includes some of the topic's main aspects but misses others, resulting in an incomplete portrayal.                                                                      \\
4:~Acceptable breadth; covers most main aspects, though it may stray into minor unnecessary details or overlook some relevant points.                                                        \\
5:~Good coverage; achieves broad coverage of the topic, hitting on all major points with minimal extraneous information.                                                                     \\
6:~Comprehensive; provides thorough coverage of all significant aspects of the topic, with a well-balanced focus.                                                                            \\
7:~Exemplary in breadth; delivers outstanding coverage, thoroughly detailing all crucial aspects of the topic without including irrelevant information.                                      \\ 
\midrule
\multicolumn{1}{c}{\textbf{Verifiability}}                                                                                                                                                   \\ 
\midrule
1:~No supporting evidence; claims are unsubstantiated.                                                                                                                                       \\
2:~Rarely supported with evidence; many claims are unsubstantiated.                                                                                                                          \\
3:~Inconsistently verified; some claims are supported; evidence is occasionally provided.                                                                                                    \\
4:~Generally verified; claims are usually supported with evidence; however, there might be a few instances where verification is lacking                                                     \\
5:~Well-supported; claims are very well supported with credible evidence, and instances of unsupported claims are rare.                                                                      \\
6:~Very well-supported; almost every claim is substantiated with credible evidence, showing a high level of thorough verification.                                                           \\
\makecell[l]{7:~Exemplary verification; each claim is supported by robust, credible evidence from authoritative sources, reflecting strict adherence to the no \\original research policy.}                  \\
\bottomrule
\end{tabular}
}
\caption{Scoring rubrics on a 1-7 scale for human evaluation.}
\label{table:human_eval_rubric}
\end{table*}

\section{Error Analysis}
\label{appendix:error_analysis}

\begin{table*}
\centering
\resizebox{\textwidth}{!}{%
\begin{tabular}{ccl} 
\toprule
\textbf{Issue}                                    & \textbf{Mentioned Time}      & \multicolumn{1}{c}{\textbf{Example Comments}}                                                                                                                                                                                                                                                                                                                                \\ 
\midrule

\multirow{11}{*}{\makecell{Use of emotional words,\\ unneutral}}            & \multirow{11}{*}{12} & \makecell[l]{The word ``significant'' is used 17 times in this article. Vague and unsupported claims are\\ made about broader political importance and ``pivotal role[s]'', and is unencyclopedic. \\(comment on article \textit{Lahaina, Hawaii})}                                                                                                                               \\ 
\cmidrule{3-3}
                                                             &                     & \makecell[l]{{[}...] but they still have not fixed the issue of neutral point of view. It is also evident in this\\ article that the writer's standpoint is biased towards Taylor Swift. Other than that, it did \\a good job at summarizing key points and putting depth into this. \\(comment on article \textit{Speak Now (Taylor's Version)})}                              \\ 
\cmidrule{3-3}
                                                             &                     & \makecell[l]{``The film was also featured in an art and film festival hosted by The California Endowment,\\ highlighting the power of stories in reshaping narratives about communities.'' Yes, technically\\ the source says that, but it's a stretch to say in  Wikipedia voice and just sounds like \\non-neutral, promotional prose. (comment on article \textit{Gehraiyaan})}  \\ 
\midrule

\multirow{6}{*}{\makecell{Red herring fallacy, \\associating unrelated sources}} & \multirow{6}{*}{11} & \makecell[l]{Polling from America shouldn't be included and links to climate change shouldn't be\\ made unless explicitly connected by the source. (comment on article \textit{Typhoon Hinnamnor})}                                                                                                                                                                         \\ 
\cmidrule{3-3}
                                                             &                     & \makecell[l]{Sourcing seems mostly fine, though some aren't directly related (Ex. 39,40). \\(comment on article \textit{Gehraiyaan})}                                                                                                                                                                                                                                       \\ 
\cmidrule{3-3}
                                                             &                     & \makecell[l]{Here is a lengthy digression about KISS, not necessary because the article on the band \\should be linked to. (comment \textit{on article 2022 AFL Grand Final})}                                                                                                                                                                                              \\ 

\midrule
\multirow{5}{*}{Missing important information}               & \multirow{5}{*}{6}  & \makecell[l]{``One study, conducted by Sinéad Griffin, a physicist at the Lawrence Berkeley National\\ Laboratory, provided some analysis of LK-99's abilities using supercomputer simulations[20].''\\ This is not enough information  about the analysis, which would have been very useful in the \\article. (comment on article \textit{LK-99})}                               \\ 
\cmidrule{3-3}
                                                             &                     & \makecell[l]{Although the earthquake's immediate aftermath and response are adequately covered, there\\ could be more about the long-term socioeconomic impact and recovery processes. \\(comment on article \textit{2022 West Java earthquake})}                                                                                                                             \\ 
\midrule
\multirow{4}{*}{\makecell{Improper handling of\\ time-sensitive information}}  & \multirow{4}{*}{5}  & \makecell[l]{Words like ``now'' should be avoided in Wikipedia articles to prevent them from becoming\\ dated and phrases such as, ``as of December 2023'' should be used instead. \\(comment on article \textit{Cyclone Batsirai})}                                                                                                                                              \\ 
\cmidrule{3-3}
                                                             &                     & \makecell[l]{``as of December 13'' doesn't specify a year, and is old information \\(comment on article \textit{2022 West Java earthquake})}                                                                                                                                                                                                                                  \\ 
\midrule
\multirow{5}{*}{Section organization problem}                & \multirow{5}{*}{5}  & \makecell[l]{too many subsections in the ``Recovery and Rehabilitation'' section\\ (comment on article \textit{2022 West Java earthquake})}                                                                                                                                                                                                                                   \\ 
\cmidrule{3-3}
                                                             &                     & \makecell[l]{I do not like how the article is organized, with too many headers cluttering the article,\\ making it not as readable. Other than that, the AI did great work on the piece. \\(comment on article \textit{2022 Crimean Bridge explosion})}                                                                                                                       \\
\bottomrule
\end{tabular}
}
\caption{Summary of major issues found on articles produced by \system.}
\label{table:comment_summary}
\end{table*}

While articles produced by \system are preferred by both automatic metrics and human evaluation, experienced editors still identified multiple problems with the machine-generated articles. We analyze the free-form comments and summarize the major issues in \reftab{table:comment_summary}.

The primary issue raised is that the generated articles often contain emotional language and lack neutrality, primarily due to the source material. \system currently retrieves grounding sources from the Internet which is not neutral and contains considerable promotional content on its own. Addressing this bias in the pre-writing stage represents a valuable direction for future research. Another major issue is the red herring fallacy or the over-association of unrelated facts. Addressing this challenge calls for high-level sensemaking rather than mere fact-level verification.

\onecolumn
\renewcommand{\arraystretch}{1.5} %
\begin{longtable}{p{\textwidth}}
\toprule
\texttt{\small{\textbf{Taylor Hawkins}}} \\
\texttt{\small{Oliver Taylor Hawkins (1972-2022) was an American musician, best known as the drummer for the rock band Foo Fighters[1]. Born in Fort Walton, Texas, Hawkins' love for music was ignited at a young age, particularly after watching a Queen concert in 1982[2][3][5]. He kick-started his professional career as the drummer for Alanis Morissette’s band during her world tour for the hit album 'Jagged Little Pill'[8][9]. His talents were recognized by Dave Grohl, who invited him to join the Foo Fighters in 1997, marking the beginning of his impactful tenure with the band[7][8].}} \\
\texttt{\small{Hawkins was celebrated for his versatile drumming style, drawing influence from renowned drummers like Roger Taylor, Neil Peart, Phil Collins, Alex Van Halen, and Stewart Copeland[14]. His performances, marked by a unique energy and aggressive style of play, earned him recognition as one of the top rock drummers of his era[15]. Apart from his role in the Foo Fighters, Hawkins' passion for music saw him involved in numerous side projects and collaborations, cementing his place in the world of rock music[10].}} \\
\texttt{\small{Outside of his professional life, Hawkins was known for his resilience and dedication to his family. Despite personal struggles, including a near-fatal drug overdose in 2001, Hawkins remained committed to his musical career[4][9]. His legacy continues to inspire musicians and fans alike, as his contributions to rock music, coupled with his indomitable spirit, made him an unforgettable icon in the music industry[13].}} \\
\texttt{\small{Hawkins' sudden death in 2022 while on tour in Bogotá, Colombia, sent shockwaves through the music world[34]. Tributes poured in from around the globe, reflecting the respect and admiration Hawkins had garnered during his lifetime[21][31]. His life and career were honored at a star-studded tribute concert in Los Angeles, attesting to the enduring impact of his music[22].}} \\

\vspace{1pt}
\texttt{\small{\textbf{\# Early Life and Background}}} \\
\texttt{\small{Oliver Taylor Hawkins, known as Taylor Hawkins, was born and raised in Fort Walton, Texas[3]. His family moved to Laguna Beach, California when he was four years old[3]. He has two younger siblings, a brother named Jason, and a sister named Heather[3]. As a child, Hawkins was particularly influenced by his paternal grandmother, Josie Hawkins, who had grown up during the Great Depression and lived in Jackson, Mississippi[1].}} \\
\texttt{\small{During his high school days at Laguna Beach High School, from where he graduated in 1990, he became friends with Jon Davison, who later became the lead vocalist of the band Yes[2][3]. His interest in music was nurtured from an early age, particularly after watching a Queen concert in 1982 which inspired him to learn to play the drums[2][5]. He noted that music was a constant presence in his family home[5].}} \\
\texttt{\small{Despite facing certain hardships during his upbringing, including his mother's struggles with "demons", Hawkins pursued his musical ambitions[4]. He credits his older sister Heather for taking care of the family during difficult times[4].}} \\
\texttt{\small{His first major musical experience came from playing drums for Alanis Morissette's album, Jagged Little Pill, and accompanying her on the subsequent tour[3]. This marked the beginning of his professional career in the music industry.}} \\

\vspace{1pt}
\texttt{\small{\textbf{\# Career}}} \\
\texttt{\small{Taylor Hawkins began his professional music career playing in Alanis Morissette’s band during her 18-month world tour in support of the hit album 'Jagged Little Pill' from 1995 to 1997[8][9]. His performances not only in the tour but also in the music videos for “You Oughta Know”, “All I Really Want” and “You Learn” introduced him to the world of rock music and ultimately led to his meeting with Dave Grohl[8]. Throughout this time, Hawkins contributed significantly to the band's sound and performance, transforming the songs from their original drum loop format to a rock-band vibe that resonated with audiences[1][7].}} \\
\texttt{\small{In 1997, Hawkins was asked by Grohl to join the Foo Fighters, an invitation that he readily accepted[7][8]. At the time, Grohl thought it was a long shot to recruit Hawkins given that Morissette was at the height of her career, but Hawkins' desire to be a part of a rock band compelled him to make the move[7]. This marked the beginning of Hawkins' tenure as the drummer of the Foo Fighters, a role that he would play until his passing[6][9].}} \\
\texttt{\small{Apart from his work with Morissette and the Foo Fighters, Hawkins had an array of other musical experiences[10]. He drummed for Sass Jordan before joining Morissette’s touring band[10]. He was part of an ad hoc drum supergroup called SOS Allstars and filled the void for Coheed and Cambria’s 2007 album after their drummer Josh Eppard left the group[10]. In addition, Hawkins formed his own side project, the Coattail Riders, in 2005, through which he recorded his own music and took the project on the road, performing in small clubs despite the Foo Fighters' arena-status[7]. His son, Shane Hawkins, has since taken on his father's legacy, joining the Foo Fighters for a performance during the Boston Calling Music Festival in 2023[6].}}\\

\vspace{1pt}
\texttt{\small{\textbf{\# Musical Style and Influences}}} \\
\texttt{\small{Taylor Hawkins was a profound drummer, with his musical style and influences spreading across a wide array of rock genres[11]. Known for his passionate fandom of groups that came before him, Hawkins regularly expressed his admiration for bands like Rush, Genesis, and the Police, all of which featured some of the greatest drummers in rock history like Neil Peart, Phil Collins, and Stewart Copeland[11].}} \\
\texttt{\small{He was heavily influenced by his love for classic rock, as evidenced by his performances, where he covered songs from bands like Van Halen[11].}} \\
\texttt{\small{Hawkins drew influences from a variety of drumming styles, developing a signature style inspired by greats like Roger Taylor, Neil Peart, Phil Collins, Alex Van Halen, and Stewart Copeland[14]. This distinctive style and influence extended to his drum kit, which incorporated elements like rototoms and concert toms[14].}} \\
\texttt{\small{Beyond his influences, Hawkins had a unique energy that made him stand out as a drummer. His performances were recognized for their power, and he was known for his enthusiastic and aggressive style of play[15]. This earned him recognition as one of the top rock drummers of his time, with his passion for music living on through his performances[14].}} \\
\texttt{\small{Through his career, Hawkins left an indelible mark on rock music, through his distinct style, passion, and contributions to the music industry[13]. His love for music and dedication to his craft made him an unforgettable icon in the world of rock music[13].}} \\

\vspace{1pt}
\texttt{\small{\textbf{\# Personal Life}}} \\
\texttt{\small{Taylor Hawkins married Alison Hawkins, an American celebrity and entrepreneur, in 2005[18]. The couple had three children, Oliver, Annabelle, and Everleigh[19]. Hawkins' commitment to his family was evident; in fact, he even wrote a song for his middle child, Annabelle[9].}} \\
\texttt{\small{In his personal life, Hawkins had also struggled with drug use, which nearly claimed his life in a 2001 overdose[9][7][4]. However, he managed to overcome this challenge, and later expressed gratitude for the experience as a lesson that allowed him to realize the destructive path he was on[7].}} \\
\texttt{\small{Outside of his main role in the Foo Fighters, Hawkins also pursued various side projects including the Birds of Satan, NHC, and Chevy Metal. His motivation for such ventures was a constant drive to create and his love for music[7]. Hawkins was also known for his unabashed fanboy nature, often vocalizing his admiration for fellow musicians and his heroes[7].}} \\

\vspace{1pt}
\texttt{\small{\textbf{\# Legacy and Impact}}} \\
\texttt{\small{Taylor Hawkins was known for his raw and authentic drumming style, described as "courageous, damaged and unflinchingly authentic"[20]. His work with the Foo Fighters, as well as his various collaborations and side projects, made him a celebrated figure in rock ‘n’ roll[10].}} \\
\texttt{\small{Hawkins' death in 2022 was met with heartfelt tributes from colleagues and fans around the world. Notable tributes came from rock legends like Roger Taylor of Queen, who considered Hawkins as a kind, brilliant man and an inspirational mentor, likening his death to "losing a younger favourite brother"[21]. Similarly, Led Zeppelin’s Jimmy Page admired his technique, energy and spirited enthusiasm[21].}} \\
\texttt{\small{An LA tribute concert held in his honor included guest drummers like Lars Ulrich of Metallica, Travis Barker of blink-182, and Brad Wilk of Rage Against the Machine. Singers like Miley Cyrus and Alanis Morissette also performed at the concert[22].}} \\
\texttt{\small{Apart from his music, Taylor Hawkins also contributed to charities Music Support and MusiCares, both of which were chosen by the Hawkins family[23]. He had received numerous accolades throughout his career, including 27 Grammy nominations, of which he won 14[2]. In 2021, the Foo Fighters were inducted into the Rock and Roll Hall of Fame[9].}} \\

\vspace{1pt}
\texttt{\small{\textbf{\# Discography}}} \\
\texttt{\small{Taylor Hawkins also led a notable music career through his own side projects and collaborations[10]. Aside from his work with the Foo Fighters, Hawkins formed and fronted the band Taylor Hawkins \& The Coattail Riders, a project which originated from jamming sessions with his friend Drew Hester[10].}} \\
\vspace{1pt}
\texttt{\small{\textbf{\#\#\# Taylor Hawkins \& The Coattail Riders}}} \\
\texttt{\small{Taylor Hawkins \& The Coattail Riders, a band formed in 2004, have released three albums and their music spans genres including Hard Rock, Art Rock, and Alternative Rock[24][25][26]. The band grew from an initial casual jamming session, gradually evolving into a more formal arrangement that led to the production of record albums. Notably, these albums featured guest appearances by renowned musicians such as Dave Grohl, Queen's Brian May and Roger Taylor, The Cars' Elliot Easton, Perry Farrell, and Jon Davison, who is a school friend of Hawkins'[10].}} \\
\vspace{1pt}
\texttt{\small{\textbf{\#\#\# Red Light Fever}}} \\
\texttt{\small{Red Light Fever, released on April 19, 2010, was the band's first album[29][30]. Prior to its release, Hawkins revealed in an interview that the album had completed the recording and production stages, but its title and release date were yet to be determined[29]. Red Light Fever was recorded at the Foo Fighters' Studio 606 in California and featured guest musicians such as Brian May and Roger Taylor of Queen, Dave Grohl of Foo Fighters, and Elliot Easton of The Cars[29][30].}} \\
\vspace{1pt}
\texttt{\small{\textbf{\#\# Get the Money}}} \\
\texttt{\small{Get the Money, the third album from Taylor Hawkins \& The Coattail Riders, was released on November 8, 2019[29]. The album's first single, "Crossed the Line", released on October 15, 2019, featured Dave Grohl and Jon Davison, the frontman of Yes[29]. The music video for the single "I Really Blew It" also featured appearances from Grohl and Perry Farrell[29].}} \\

\vspace{1pt}
\texttt{\small{\textbf{\# Collaborations and Guest Appearances}}} \\
\texttt{\small{Throughout his career, Taylor Hawkins collaborated with various prominent artists and bands. The Coattail Riders' albums notably featured appearances from luminaries such as Brian May and Roger Taylor of Queen, Chrissie Hynde, Nancy Wilson of Heart, Sex Pistol Steve Jones and James Gang's Joe Walsh[28]. Hawkins also fronted another group, The Birds of Satan, which evolved from his heavy rock covers band, Chevy Metal[28].}} \\
\texttt{\small{Despite his diverse musical engagements, Hawkins always maintained a close allegiance with the Foo Fighters, which remained the center of his music life[7][28].}} \\

\vspace{1pt}
\texttt{\small{\textbf{\# Tragic Passing}}} \\
\texttt{\small{Taylor Hawkins, the esteemed drummer of the alt-rock band Foo Fighters, passed away suddenly on March 25, 2022, while on tour with his band in Bogotá, Colombia[34]. The official cause of death was cardiac arrest, though inquiries were raised concerning the presence of drugs in his system and their potential contribution to his death[33][34]. On the night of his passing, paramedics were called to the Four Seasons hotel in Bogotá due to reports of chest pain from an unnamed guest, later revealed to be Hawkins[34]. Unfortunately, resuscitation efforts were unsuccessful, and Hawkins was declared dead at the scene[34].}} \\
\texttt{\small{The news of Hawkins' sudden demise was announced on the morning of March 25th, 2022, which left the music world in shock[32]. The band confirmed the news with a short statement, expressing their devastation at the loss of Hawkins, whose "musical spirit and infectious laughter" would live on forever[32].}} \\
\texttt{\small{As a result of Hawkins' untimely passing, the band canceled their ongoing South American tour[33]. The festival stage at the Estéreo Picnic Festival, where the Foo Fighters were scheduled to perform that night, was transformed into a candlelight vigil in memory of Hawkins[33].}} \\
\vspace{1pt}
\texttt{\small{\textbf{\#\# Tributes and Remembrances}}} \\
\texttt{\small{In the wake of Hawkins' death, tributes from fans and colleagues alike poured in from around the world[21][31]. Among the many paying their respects were legendary rock and roll musicians like Roger Taylor, the drummer of Queen, who Hawkins credited with inspiring his own career behind the drum set[21]. In heartfelt social media posts, Taylor described Hawkins as an "inspirational mentor" and a "kind brilliant man"[21], while Led Zeppelin's Jimmy Page reminisced about sharing the stage with Hawkins and praised his "technique, energy and spirited enthusiasm"[21].}} \\
\texttt{\small{There were also numerous onstage tributes to Hawkins. Notably, Miley Cyrus expressed her grief and sent peaceful wishes to the Foo Fighters and the Hawkins family during a performance at Lollapalooza[31]. Similarly, Liam Gallagher of Oasis dedicated one of the band's biggest hits to Hawkins during a concert at the Royal Albert Hall in London[31].}} \\
\texttt{\small{Fans gathered outside the hotel where Hawkins died, lighting candles, leaving flowers, and singing the band's songs in his honor[31].}} \\
\texttt{\small{Hawkins' life and career were celebrated in a star-studded tribute concert in Los Angeles, which saw performances from over 50 musicians, including his former bands and colleagues from Def Leppard, Queen, and Foo Fighters[22].}} \\

\bottomrule
\caption{\system's generated article for ``Taylor Hawkins''. ``\#'', ``\#\#'' indicate the section title and subsection title respectively. Numbers in brackets indicate the cited references.}
\label{table:example}
\end{longtable}

\end{document}